\documentclass[10pt,journal,compsoc]{IEEEtran} 
\ifCLASSOPTIONcompsoc
  \usepackage[nocompress]{cite}
\else
  \usepackage{cite}
\fi
 \pdfoutput=1
\usepackage{array}
\usepackage{hyperref}

\usepackage{amsmath}
\usepackage{amssymb}
\usepackage{amsthm}
\usepackage[linesnumbered,ruled,vlined]{algorithm2e}
\SetAlCapNameFnt{\small}
\usepackage{algpseudocode}
\usepackage{verbatim}
\usepackage{multirow}
\usepackage{arydshln}
\usepackage{bm}
\usepackage{mathtools}
\usepackage{dsfont}
\usepackage{latexsym} 
\usepackage{comment}
\usepackage{xcolor}
\usepackage{graphicx}
\usepackage{enumitem}
\usepackage{diagbox}
\usepackage{hyphenat}
\usepackage[font={footnotesize,sf},subrefformat=simple,labelformat=simple,justification=centering]{subcaption}
\usepackage{indentfirst}
\usepackage{setspace}
\usepackage{booktabs}
\usepackage[numbers]{natbib}
\usepackage[capitalize]{cleveref}
\usepackage[font={footnotesize,sf},justification=justified,singlelinecheck=false]{caption}

\graphicspath{{./figures/}}

\DeclareMathOperator*{\argmax}{arg\,max}

\DeclarePairedDelimiter\abs{\lvert}{\rvert}%
\DeclarePairedDelimiter\norm{\lVert}{\rVert}%

\makeatletter
\let\oldabs\abs
\def\abs{\@ifstar{\oldabs}{\oldabs*}}
\let\oldnorm\norm
\def\norm{\@ifstar{\oldnorm}{\oldnorm*}}
\makeatother

\newcolumntype{L}[1]{>{\raggedright\let\newline\\\arraybackslash\hspace{0pt}}m{#1}}
\newcolumntype{C}[1]{>{\centering\let\newline\\\arraybackslash\hspace{0pt}}m{#1}}
\newcolumntype{R}[1]{>{\raggedleft\let\newline\\\arraybackslash\hspace{0pt}}m{#1}}

\begin{document}

\title{$\mathcal{X}$\hyp{}Metric: An N\hyp{}Dimensional Information\hyp{}Theoretic Framework for Groupwise Registration and Deep Combined Computing}
\author{Xinzhe~Luo and Xiahai~Zhuang*
\IEEEcompsocitemizethanks{\IEEEcompsocthanksitem X. Luo and X. Zhuang are with the School of Data Science, Fudan University, 220 Handan Road, Shanghai, 200433, China. 
\IEEEcompsocthanksitem *X. Zhuang is the corresponding author (\url{www.sdspeople.fudan.edu.cn/zhuangxiahai/}).
\IEEEcompsocthanksitem Code will be available from \url{https://zmiclab.github.io/projects.html}.
}
\thanks{Manuscript received ..; revised ..}
}

\IEEEtitleabstractindextext{%
\begin{abstract}

This paper presents a generic probabilistic framework for estimating the statistical dependency and finding the anatomical correspondences among an arbitrary number of medical images.
The method builds on a novel formulation of the $N$-dimensional joint intensity distribution by representing the common anatomy as latent variables and estimating the appearance model with nonparametric estimators.
Through connection to maximum likelihood and the expectation-maximization algorithm, an information\hyp{}theoretic metric called $\mathcal{X}$-metric and a co-registration algorithm named $\mathcal{X}$-CoReg are induced, allowing groupwise registration of the $N$ observed images with computational complexity of $\mathcal{O}(N)$.
Moreover, the method naturally extends for a weakly-supervised scenario where anatomical labels of certain images are provided.
This leads to a combined\hyp{}computing framework implemented with deep learning, which performs registration and segmentation simultaneously and collaboratively in an end-to-end fashion.
Extensive experiments were conducted to demonstrate the versatility and applicability of our model, including multimodal groupwise registration, motion correction for dynamic contrast enhanced magnetic resonance images, and deep combined computing for multimodal medical images.
Results show the superiority of our method in various applications in terms of both accuracy and efficiency, highlighting the advantage of the proposed representation of the imaging process.
\end{abstract}

}

\maketitle
\IEEEdisplaynontitleabstractindextext
\IEEEpeerreviewmaketitle

\IEEEraisesectionheading{\section{Introduction}\label{sec:intro}}
\IEEEPARstart{M}{edical} images are usually complementary yet inherently correlated through their underlying common anatomy.
Mutual information (MI), an information\hyp{}theoretic metric measuring the statistical dependency between two images, has been particularly successful in medical image registration \cite{journal/IJCV/viola1997, journal/tmi/maes1997}.
{However, generalizing MI to $N$ images
for $N\gg 2$ can be challenging due to the curse of dimensionality.
To devise computational tractable solutions, \citet{journal/tpami/learned2005} proposed the pixel/voxel-wise stacked entropies, which works on \emph{i.i.d.} entries of each intensity vector at individual spatial locations;
\citet{journal/tpami/wachinger2012} worked on the pairwise estimates of MI, which accumulates MI's over all image pairs;
\citet{conference/miccai/bhatia2007} and \citet{journal/media/polfliet2018} resorted to template-based methods, which rely on an informative grey-valued template generated during the registration process.}


In this paper, we choose to pay close attention to the connection between information-theoretic metrics and their interpretation from maximum likelihood.
Similar approaches were taken in \cite{journal/IMA/roche2000,journal/media/sedghi2021} for the special cases of pairwise registration or \emph{i.i.d.} intensity vectors. 
Unlike their methods, our approach is to consider modelling the joint intensity distribution directly by introducing latent variables and using nonparametric estimators.
In this manner, we derive an $N$\hyp{}dimensional information\hyp{}theoretic framework, composed of a groupwise similarity metric and a generic co\hyp{}registration algorithm with computational complexity of $\mathcal{O}(N)$.
As the terminology has yet to be precisely defined, they are referred to as the $\mathcal{X}$-metric and the $\mathcal{X}$-CoReg algorithm, respectively.

The core of the $\mathcal{X}$-metric and {the} $\mathcal{X}$-CoReg is the explicit modelling of the common anatomy as categorical latent variables, along with a generative appearance model using nonparametric estimators.
These major components yield an optimization scheme inspired by the expectation-maximization algorithm that alternates between:  
1) estimating the common-space parameters, including the spatial distribution and prior proportions of the common anatomy (\cref{sec:update_template}); 
2) identifying the spatial correspondences among the observed images (\cref{sec:update_transform}).
{Besides}, the approach goes beyond registration;
it actually integrates registration with segmentation, leading to a combined\hyp{}computing framework that performs the two tasks simultaneously and collaboratively \cite{conference/miccai/xiaohua2004,conference/ipmi/xiahua2005,journal/NI/pohl2006,conference/miccai/bhatia2007/groupwise,journal/tpami/Zhuang2019}.

To demonstrate its versatility, we validate $\mathcal{X}$-CoReg on groupwise registration, including joint alignment for multimodal images of the brain and the most challenging task on motion correction for first-pass cardiac perfusion sequences.
Furthermore, to leverage the capability of deep learning (DL) techniques, we extend the proposed algorithm to a DL-integrated framework, so that end-to-end combined computing is realized on multimodal medical images.

Our contributions can be summarized as follows:
\begin{itemize}[leftmargin=*]
  \item We propose an information-theoretic and computationally tractable co-registration framework, which highlights:
  \begin{itemize}
    \renewcommand\labelitemi{-}
    \item The $\mathcal{X}$-metric that is intended to measure the statistical dependency among an arbitrary number of images, based on a novel formulation of the $N$-dimensional joint intensity distribution and the imaging process. 
    \item The $\mathcal{X}$-CoReg algorithm that features a generic yet efficient co-registration procedure. 
    \item The proposed framework and optimization scheme are theoretically grounded from the maximum-likelihood perspective. 
  \end{itemize}
  \item We extensively evaluate the applicability of the proposed algorithm on multimodal groupwise registration and motion estimation for dynamic contrast-enhanced images, and find its superiority over previous methods.
  \item We demonstrate the potential of the algorithm in a deep learning context, where simultaneous registration and segmentation is achieved on multimodal medical images in an end-to-end fashion.
\end{itemize}

\begin{figure}
  \centering
  \includegraphics[width=\linewidth]{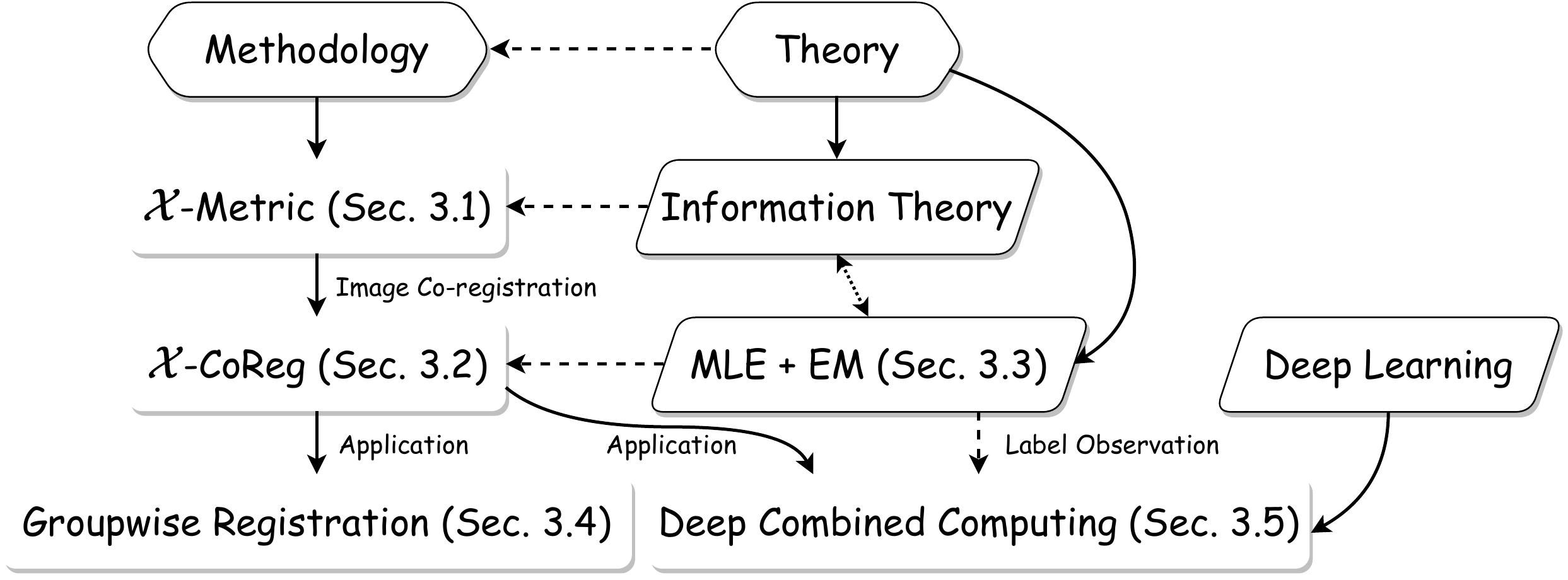}
  \caption{{Roadmap of the proposed framework.}}
  \label{fig:roadmap}
\end{figure}

The remainder of the paper is organized as follows.
\cref{sec:ML_perspective} provides an overview of the past studies on groupwise registration and combined computing, with special emphasis on the connection between information\hyp{}theoretic approaches and the maximum likelihood principle.
{
\cref{sec:ITCR} describes the {proposed} $N$-dimensional information-theoretic framework---the $\mathcal{X}$-metric (\cref{sec:X-metric}) and {the} $\mathcal{X}$-CoReg (\cref{sec:X-CoReg}), whose rationale are established from the generic image generative model and maximum likelihood (\cref{sec:MLE}).
We also discuss the application of the proposed algorithm to groupwise registration (\cref{sec:GR}) and an extended framework that incorporates observed labels, yielding its application to deep combined computing (\cref{sec:DCC}).
\cref{sec:experiment} presents experimental setups and evaluation results of {our method} on applications to multimodal groupwise registration (\crefrange{sec:BrainWeb}{sec:RIRE}), spatiotemporal motion estimation (\cref{sec:MoCo}), and deep combined computing (\cref{sec:MSCMR}).
\cref{sec:discussion} discusses implications of the proposed framework and concludes the study.
\cref{fig:roadmap} presents the roadmap of the proposed framework.
}

\begin{figure}
  \centering
  \begin{subfigure}[b]{0.9\linewidth}
    \centering
    \includegraphics[width=\textwidth]{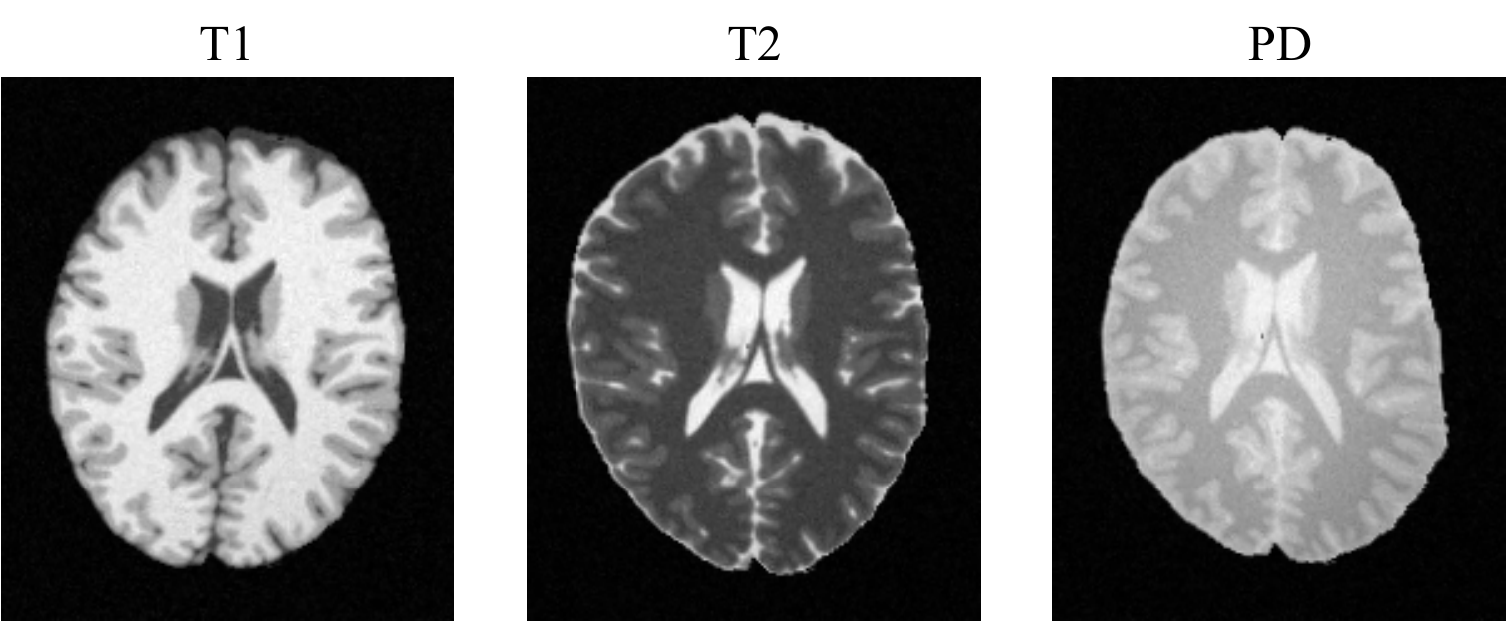}
    \caption{Misaligned multi-sequence MR images from BrainWeb.}
  \end{subfigure}
  \\
  \vspace{0.2cm}
  \captionsetup[subfigure]{justification=justified}
  \begin{subfigure}[b]{0.46\linewidth}
    \centering
    \includegraphics[width=\textwidth]{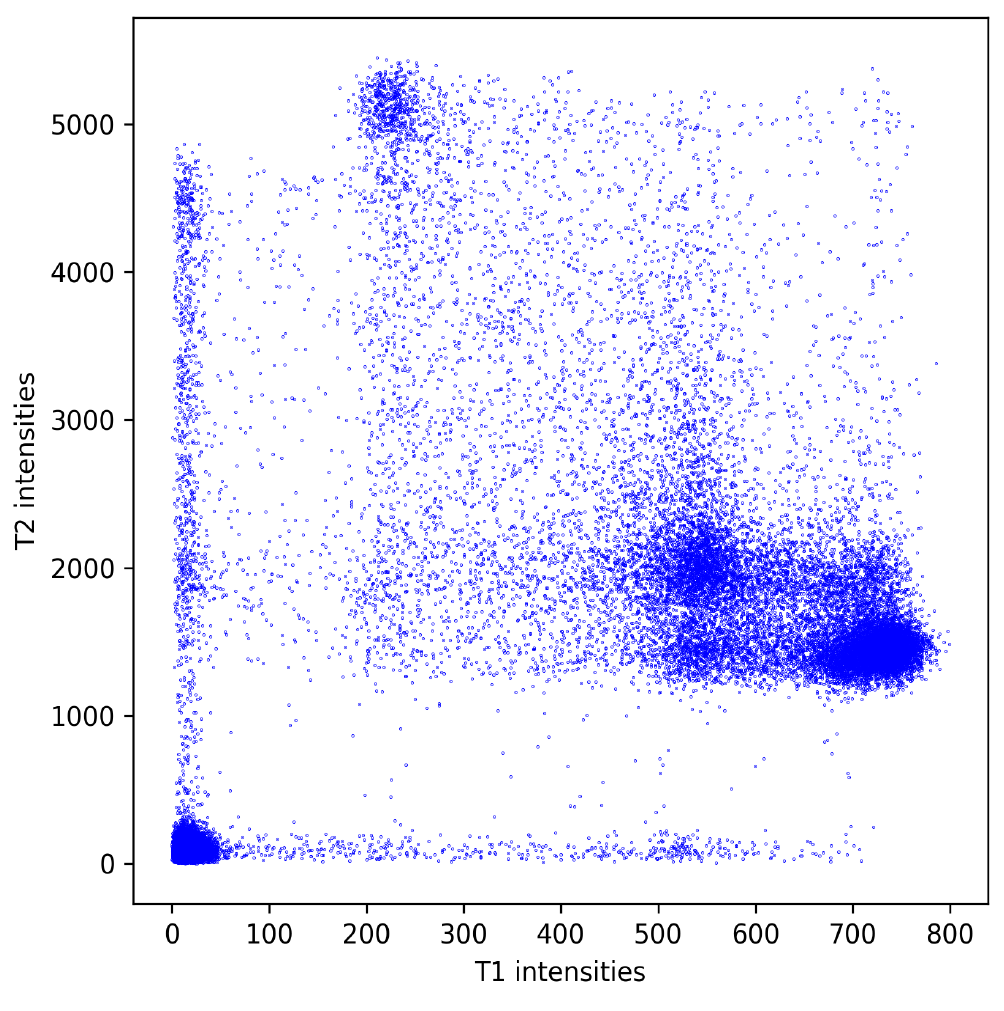}
    \caption{{Scatter plot of intensity pairs. 
    Each point indicates one intensity pair from the T1 and T2 images at the same spatial location.}}
  \end{subfigure}
  \hfill
  \begin{subfigure}[b]{0.5\linewidth}
    \centering
    \includegraphics[width=\textwidth]{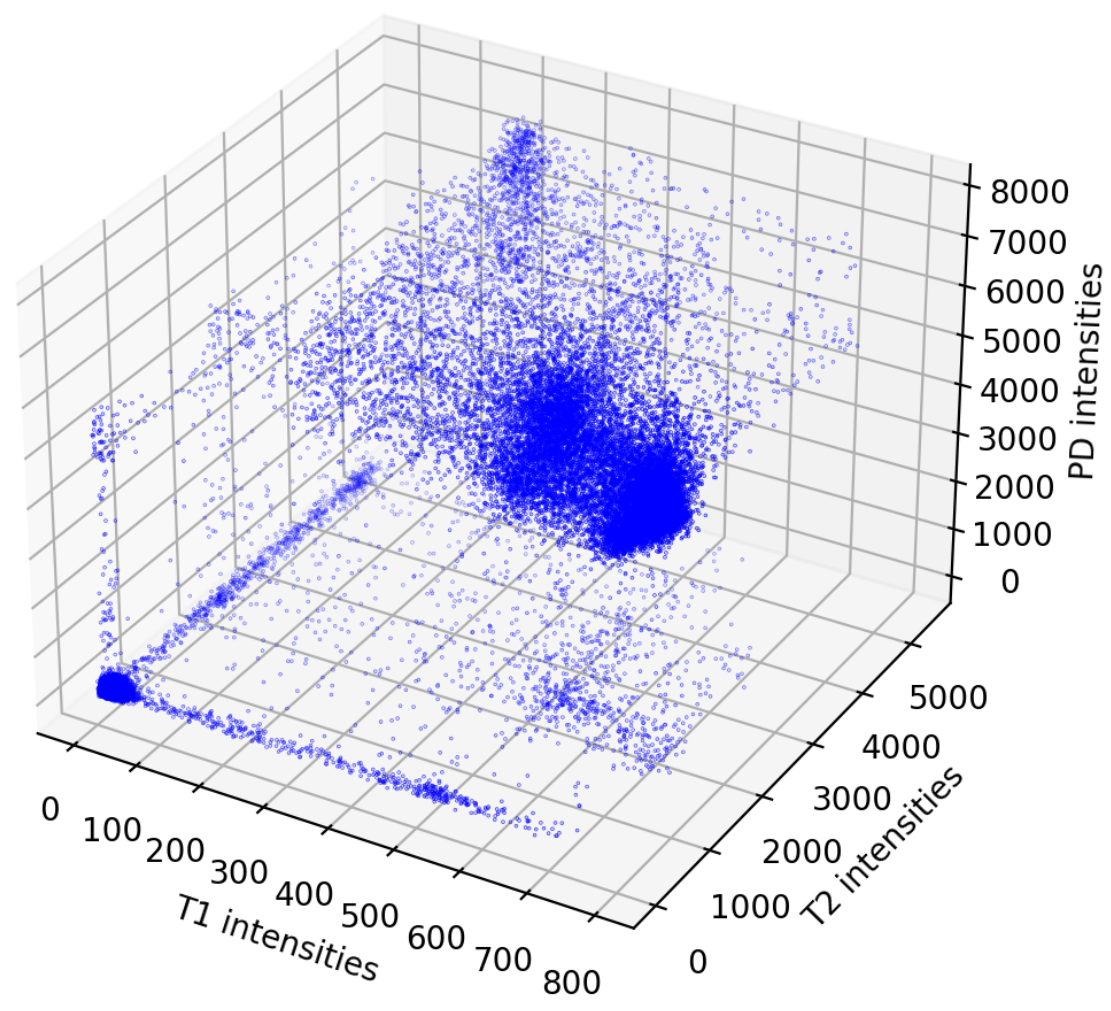}
    \caption{{Scatter plot of intensity triples.
    Each point indicates one intensity triple from the T1, T2 and PD images at the same spatial location.}}
  \end{subfigure}
  \caption{The joint intensity scatter plot of the multi-sequence MR images from the BrainWeb dataset with simulated misalignments.
  {The average number of samples in a single bin of the joint intensity space (JIS), calculated as $|\Omega| / L^N$ where $L$ is the number of discrete intensity levels of an image and $|\Omega|$ is the constant cardinality of the common-space spatial samples, will reduce exponentially as the dimension $N$ of the joint intensity space increases.}}
  \label{fig:scatter_plot}
\end{figure}

\begin{table*}[t]
  \small
  \centering
  \caption{Definition of the mathematical symbols used in this paper.}
  
  \begin{tabular}{p{0.12\columnwidth}p{0.8\columnwidth}|p{0.12\columnwidth}p{0.8\columnwidth}}
    \toprule
    Symbol & Description & Symbol & Description \\
    \midrule
    $N$ & the number of observed images & 
    $j$ & index of the observed image, $j\in\{1,\dots,N\}$ \\

    $K$ & the number of common anatomical labels & 
    $k$ & index of the anatomical label, $k\in\{1,\dots,K\}$ \\
    
    \noalign{\vskip 0.2ex}\hdashline\noalign{\vskip 0.5ex}

    $\Omega$ & the common space with discrete Cartesian grid & $\abs*{\Omega}$ & the cardinality of the common space \\

    $\bm{Z}$ & categorical model of the common anatomy & 
    $\bm{z_x}$ & one-hot vector of the common anatomy at $\bm{x}$ \\

    $\bm{\Gamma}$ & spatial distribution of the common anatomy & 
    $\bm{\gamma_x}$ & probability vector of the common anatomy at $\bm{x}$ \\

    $\bm{\pi}$ & prior proportions of the common anatomy & 
    $\Omega_j$ & the $j$-th image space \\

    $\bm{U}$ & the observed image group $\bm{U}=\{U_j\}_{j=1}^N$ &
    $U_j$ & the $j$-th observed image, $U_j=(u_{\bm{x},j})_{\bm{x}\in\Omega_j}$ \\
    
    \noalign{\vskip 0.2ex}\hdashline\noalign{\vskip 0.5ex}

    $\mathcal{J}$ & indices of the images with given segmentation &
    $\widehat{\bm{Y}}_j$ & the predicted segmentation mask of $U_j$ \\

    $\bm{Y}_j$ & the available segmentation mask of $U_j$ &
    $\widehat{\bm{\rho}}_j$ & the predicted probability maps of $U_j$ \\

    \noalign{\vskip 0.2ex}\hdashline\noalign{\vskip 0.5ex}
    
    $\bm{\phi}$ & the set of transformations $\bm{\phi}=\{\phi_j\}_{j=1}^N$ &
    $\phi_j$ & the spatial transformation from $\Omega$ to $\Omega_j$ \\
    
    $\bm{U}[\bm{\phi}]$ & the warped image group $\bm{U}[\bm{\phi}]=\{U_j\circ\phi_j\}_{j=1}^N$ &
    $\Omega^{\bm{\phi}}$ & the overlap region given by $\bm{\phi}=\{\phi_j\}_{j=1}^N$ \\

    \noalign{\vskip 0.2ex}\hdashline\noalign{\vskip 0.5ex}
    
    {$u_{j\bm{\omega}}$} & {abbreviation for ${U}_j(\bm{\omega})$, where $\bm{\omega}\in\Omega_j$} &
    $v_{\bm{x}}$ & the template intensity at $\bm{x}$ computed by PCA \\

    $\bm{u_x^{\phi}}$ & the resampled intensity vector $\bm{u_x^{\phi}}=(u_{\bm{x},j}^{\phi_j})_{j=1}^N$ &
    $u_{\bm{x},j}^{\phi_j}$ & {abbreviation for ${U}_j\circ\phi_j(\bm{x})$, where $\bm{x}\in\Omega$} \\

    \noalign{\vskip 0.2ex}\hdashline\noalign{\vskip 0.5ex}

    $L$ & the assumed number of intensity levels  &
    $\mu_j$ & a certain intensity level of $U_j$ for $j=1,\dots,N$ \\

    $\bm{\mu}$ & the intensity level vector $\bm{\mu}=(\mu_j)_{j=1}^N$ &
    $\mu_{\bm{x},j}^{\phi_j}$ & the corresponding intensity level for $u_{\bm{x},j}^{\phi_j}$ \\
    
    \bottomrule
  \end{tabular}
  \label{tab:notation}
\end{table*}

\section{The Maximum-Likelihood Perspective}\label{sec:ML_perspective}
In this section, we review the literature on groupwise registration and combined computing from the perspective of maximum likelihood.
The two tasks reduce to pairwise registration and atlas-based segmentation respectively, when only two images are modelled.
For these two reduced problems, \citet{journal/IMA/roche2000} and \citet{journal/NI/ashburner2005} have formulated them as maximum likelihood estimation (MLE), the solution to which gives the required spatial correspondence and/or segmentation with the target image.

However, direct generalization of the likelihood function to an arbitrary number of images can be complex, as the sparsity of samples in a high-dimensional intensity space will cause the estimation of joint intensity distribution (JID) to be implausible \cite{journal/tip/spiclin2012, journal/tip/orchard2009}.
\cref{fig:scatter_plot} shows an example of 2D ($N=2$) and 3D ($N=3$) joint intensity scatter plots from a misaligned image group.
{The average number of samples in a single bin of the joint intensity space (JIS), calculated as ${\abs{\Omega}}/{L^N}$ where $L$ is the number of discrete intensity levels of an image and $\abs{\Omega}$ is the constant cardinality of the common-space spatial samples, will reduce exponentially as the dimension $N$ of the joint intensity space increases.
This leads to the curse of dimensionality, which in our context means the number of samples required to consistently estimate probability distributions grows exponentially with $N$ \cite{journal/tip/spiclin2012}.}
Hence, additional independence and model assumptions over the JID are often prescribed, to make solutions computationally feasible for high-dimensional cases.

\begin{figure*}[t]
  \centering
  \includegraphics[width=\textwidth]{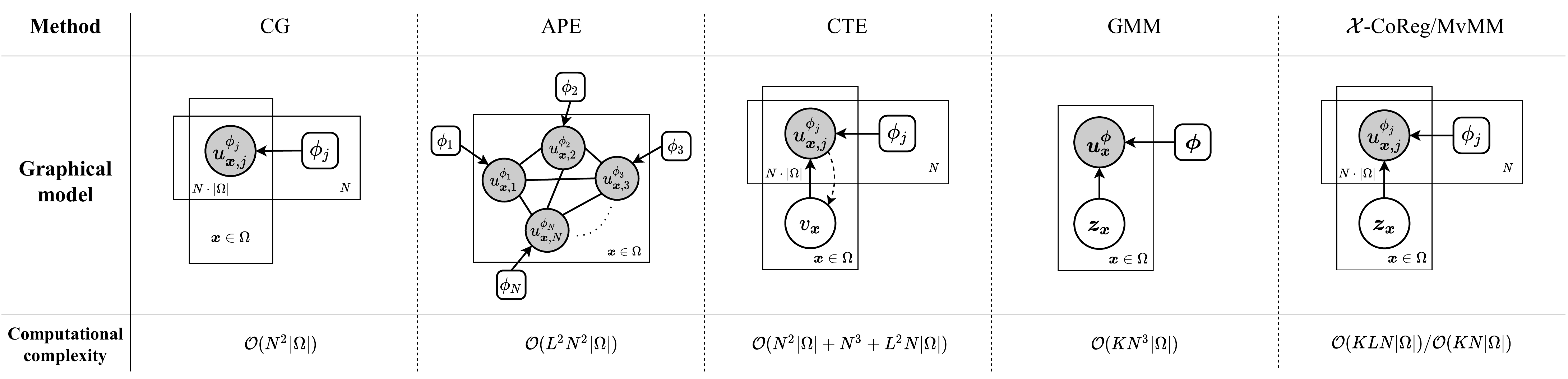}
  \caption{The graphical model and the computational complexity of the methods reviewed and investigated in this paper, where we assume the sample space is the entirety of $\Omega$. 
  Note that random variables are in circles, deterministic parameters are in rounded boxes, observed variables are shaded and boxes indicate replication. Solid arrows indicate generation while dashed ones refer to inference from principal component analysis.}
  \label{fig:graphical_model_all}
\end{figure*}

In the following, we describe the methodological details of the relevant literature.
\emph{For convenience, we list in \cref{tab:notation} the essential mathematical symbols used in the rest of this paper, and show in \cref{fig:graphical_model_all} the graphical representation of the methods reviewed and investigated in this paper.}

\subsection{Groupwise Registration}
{
  For groupwise registration, the observed image group are denoted by $\bm{U}=\{U_j\}_{j=1}^N$, where $U_j$ is the realization of some imaging process from the image space $\Omega_j$ to real intensity values.
  We can thus represent $U_j=(u_{j\bm{\omega}})_{\bm{\omega}\in\Omega_j}$, where $u_{j\bm{\omega}}\triangleq{U}_j(\bm{\omega})$ are signal intensities at \emph{i.i.d.} spatial samples $\bm{\omega}\in\Omega_j$ \cite{conference/ipmi/zollei2003}.
  Denote $\bm{u_x^{\phi}}$ as the resampled intensity vector comprising intensity values of $U_j$ at $\phi_j(\bm{x})$, \emph{i.e.} $\bm{u_x^{\phi}}=[u_{\bm{x},1}^{\phi_1},\dots,u_{\bm{x},N}^{\phi_N}]^{\intercal}$, where $u_{\bm{x},j}^{\phi_j}\triangleq{U}_j\circ\phi_j(\bm{x})$.
  The maximum-likelihood approach aims to find the optimal spatial correspondences through the MLE of a \emph{multivariate} JID indexed by the spatial transformations $\bm{\phi}$, \emph{i.e.},
  \begin{equation}
    P_{\bm{\phi}}(\bm{U})=\prod_{\bm{x}\in\Omega}P^{\bm{x}}(\bm{u_x^{\phi}}),
  \end{equation}
  which is factorized over \emph{i.i.d.} spatial samples $\bm{x}\in\Omega$.
  }
The superscript $\bm{x}$ of $P^{\bm{x}}$ indicates that the distribution for every intensity vector can be spatially variant, and will be omitted {otherwise}.
Notably, by representing the JID with a spatially invariant categorical model, we can derive the joint entropy $H(\bm{U})$ as a {groupwise similarity metric} \cite{journal/tip/spiclin2012}.

As the joint entropy can be computationally prohibitive in general when $N\gg 2$, the following studies presume extra structures with the JID, yielding more computationally tractable information-theoretic metrics:
\begin{itemize}[leftmargin=*]
  \item \emph{Congealing (CG).} 
  The CG framework \cite{conference/CVPR/miller2000,journal/tpami/learned2005} works on the \emph{i.i.d.} assumption of the entries in every intensity vector \cite{journal/media/sedghi2021}, \emph{i.e.},
  \begin{equation}
    P^{\bm{x}}(\bm{u_x^{\phi}})=\prod_{j=1}^N f_U^{\bm{x}}(u_{\bm{x},j}^{\phi_j}),
  \end{equation}
  where $f_U^{\bm{x}}(\cdot)$ is a spatially variant \emph{univariate} density.
  Therefore, the maximum likelihood principle reduces to minimum sum of pixel/voxel-wise stacked entropies, favouring transformations that force the local intensity space concentrated \cite{journal/tpami/cordero2013}.
  However, congealing with pixel stacks may suffer from an underestimated density with insufficient samples when $N$ is small.
  To mitigate this issue, \cite{journal/tpami/learned2005} also discussed the idea of pixel cylinder that considers neighbouring pixels for density estimation.
  \item \emph{Accumulated pairwise estimates (APE).}
  The APE framework \cite{journal/tpami/wachinger2012} assumes that every two images are conditionally independent given a third one.
  That is, the set of the observed images forms a fully-connected pairwise Markov random field.
  Hence, the JID is given by a Gibbs distribution
  \begin{equation}
    P(\bm{u_x^{\phi}}) \propto \prod_{j=1}^N\prod_{i>j}\psi_{ij}(u_{\bm{x},i}^{\phi_i},u_{\bm{x},j}^{\phi_j}), 
  \end{equation}
  where $\psi_{ij}(\cdot,\cdot)$ are the pairwise potentials.
  The APE measures the statistical dependency between two nodes via $\psi_{ij}(\mu_i,\mu_j)=f_{ij}(\mu_i,\mu_j)/[f_i(\mu_i)f_j(\mu_j)]$ using a categorical model, where $f_{ij}(\cdot,\cdot)$ denotes the pairwise \emph{pmf} (probability mass function) and $f_i(\cdot)$ its marginal.
  Then, the log-likelihood function devolves to the sum of all pairwise mutual information.
  Hence, the computational burden for APE scales quadratically with $N$, limiting its applicability for large image groups.
  \item \emph{Conditional template entropy (CTE).}
  Template-based groupwise registration methods are also widely adopted in the literature \cite{conference/miccai/bhatia2007, journal/PMB/melbourne2007, journal/media/metz2011,journal/media/polfliet2018}, in which the similarity metric is computed between each image and an artificial template generated during the registration process.
  Among them the conditional template entropy \cite{journal/media/polfliet2018} computes the template image as the first principal component of the observed images via principal component analysis (PCA).
  Then the sum of conditional entropies with each observed image given the template is used as the similarity metric.
  However, the principal component image may disguise clear structural information as it is computed from the linear combination of the warped images. 
  Besides, the approach is theoretically inconsistent as the (probabilistic) PCA assumes a linear Gaussian model with a latent Gaussian variable corresponding to the principal-component subspace, while the conditional entropy is derived under a categorical model \cite{journal/IMA/roche2000}.
\end{itemize}
Nevertheless, one can find that each method has certain limitations that restrict their generalizability.
To establish a general co-registration framework, we seek to construct a probabilistic generative model based solely on the primary assumption that the images are correlated through the common anatomy.
Thus, we derive the proposed $N$-dimensional information-theoretic framework, \emph{i.e.} the $\mathcal{X}$-metric and $\mathcal{X}$-CoReg, which is generic by nature.

\subsection{Combined Computing}
Beyond groupwise registration, another stream of research attempts to achieve combined computing, \emph{i.e.} combining registration with segmentation in a unified framework \cite{journal/media/lorenzen2006, conference/miccai/bhatia2007/groupwise, journal/tip/orchard2009, journal/NI/blaiotta2018, journal/tpami/Zhuang2019}. 
They typically identify the structural correspondences among a group of images by modelling the average tissue classes as latent variables.

In an early work, \citet{journal/media/lorenzen2006} chose to estimate the class posteriors of every individual image using the Gaussian mixture model (GMM).
Then, groupwise registration was realized simultaneously with finding an average atlas probability map, by minimizing the Kullback-Leibler divergence between the atlas and each class posterior.
\citet{conference/miccai/bhatia2007/groupwise} extended this notion by interleaving groupwise segmentation with groupwise registration, where the GMM parameters were optimized alternately with the registration parameters.
Nevertheless, the two methods are less principled as no explicit modelling of the JID is presented.

Thenceforth, \citet{journal/tip/orchard2009} proposed representing the JID directly with a \emph{multivariate} GMM.
Their approach alternates between density estimation and motion adjustment, corresponding to the two steps of the \emph{expectation-maximization (EM)} algorithm.
However, it has computational complexity of $\mathcal{O}(N^3)$, preventing its usage for a large image group.
\citet{journal/NI/blaiotta2018} suggested a Bayesian framework with a factorized GMM for the JID, where the intensities were assumed conditionally independent given the tissue class labelling, \emph{i.e.},
\begin{equation}\label{eq:cond_ind}
  P(\bm{U}\mid \bm{Z}) = \prod_{j=1}^N P(U_j\mid \bm{Z}).
\end{equation}
\citet{journal/tpami/Zhuang2019} sought to replace the class conditional distribution $P(U_j\mid \bm{Z})$ with another mixture of Gaussians, generating an enhanced predictive JID.
These two approaches have the advantage of computational efficiency due to the assumption of \cref{eq:cond_ind}.

{Following the same conditional independence assumption, we propose modelling the class conditional distribution with a categorical model.
Then, by leveraging the EM algorithm, we have found that the expected complete-data log-likelihood has a strong connection to the proposed information-theoretic $\mathcal{X}$\hyp{}metric.}
The performance of the induced co-registration algorithm also exceeds that of the previous GMM-based methods.

Furthermore, the proposed algorithm extends naturally to a weakly-supervised scenario where the anatomical labels of a subset of images are provided.
This feature enables a deep-learning framework that estimates segmentation and registration simultaneously and collaboratively in an end-to-end fashion, \emph{i.e.} the deep combined computing.

\section{Information-Theoretic Co-Registration}\label{sec:ITCR}
Given $N$ observed images $\bm{U}=\{U_j\}_{j=1}^N$, the purpose of co-registration is to find the spatial transformations $\bm{\phi}=\{\phi_j\}_{j=1}^N$ that aligns them into a common coordinate system $\Omega$.
This can be accomplished by an information-theoretic metric defined over $\bm{U}$.
In information theory, the statistical dependency of a set of random variables can be measured by their total correlation, \emph{i.e.} the Kullback\hyp{}Leibler (KL) divergence between the joint distribution and the product of its marginals
\begin{equation}\label{eq:total_correlation}
  \begin{aligned}
    C(\bm{U}) &\triangleq D_{\mathrm{KL}}\left[P(\bm{U})\,\Big\Vert\,\prod_{j=1}^N P(U_j)\right] \\
    &= \left[\sum_{j=1}^N H(U_j)\right] - H(\bm{U}),
  \end{aligned}
\end{equation}
where $H(\cdot)$ is the Shannon's entropy that evaluates the uncertainty of a random variable/vector.
By definition, the total correlation is {nonnegative} and is maximized if the random variables are related by bijections so that one of them determines all the others.
These properties can be exploited to achieve image registration, where the objective is to recover the spatial correspondences of two or multiple images by maximizing a given similarity metric.
For instance, mutual information (MI), as a reduced version of the total correlation when $N=2$, is widely adopted as a similarity metric for pairwise image registration \cite{journal/IJCV/viola1997,journal/tmi/maes1997,journal/tmi/pluim2003}.

However, computation of the joint entropy $H(\bm{U})$ entails the construction of a JID $P(\bm{U})$, which suffers from the curse of dimensionality for $N\gg 2$ with the widely adopted kernel density estimators \cite{journal/tip/spiclin2012}.
Yet for medical images, it is usually the case that $\bm{U}$ comprises a cohort of images from imaging processes that reflect a \emph{common} anatomy.
Let $U_j$ be the $j$-th observed image within the cohort as a realization of some stochastic imaging process over a discrete Cartesian grid ${\Omega}_j\subset\mathds{R}^d$, with $d$ the dimensionality of the sample space.
These images can be considered as transmitted through distinct image formation channels (including shape and appearance transition) from the common anatomy, also referred to as the \emph{common space} $\Omega\subset\mathds{R}^d$ \cite{journal/tpami/Zhuang2019}.
{The common space $\Omega$ is represented by a categorical random field $\bm{Z}=(\bm{z_x})_{\bm{x}\in\Omega}$, where each $\bm{z_x}=[z_{\bm{x},1},\dots,z_{\bm{x},K}]^{\intercal}$ is a one-hot vector such that $z_{\bm{x},k}=1$ if and only if the spatial location $\bm{x}$ belongs to the $k$-th tissue class.
}

\begin{figure}[t]
  \centering
  \begin{subfigure}[b]{0.34\columnwidth}
    \centering
    \includegraphics[width=\textwidth]{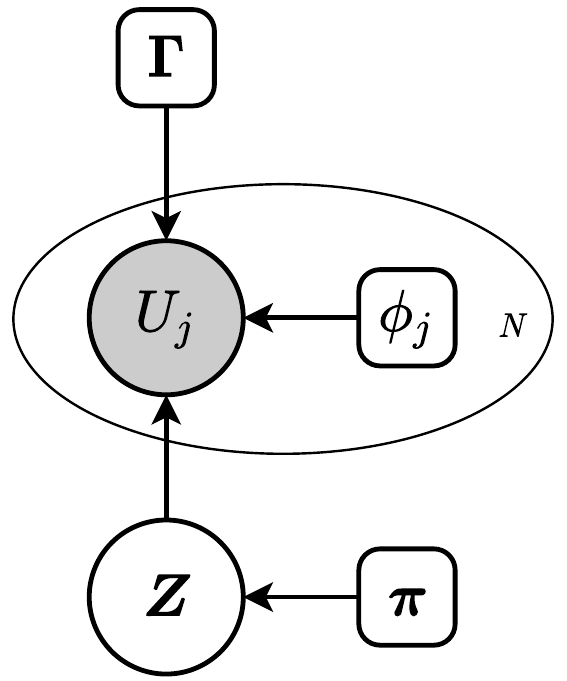}
    \caption{Generic framework.}
    \label{fig:graphical_model_a}
  \end{subfigure}
  \qquad\quad
  \begin{subfigure}[b]{0.4\columnwidth}
    \centering
    \includegraphics[width=\textwidth]{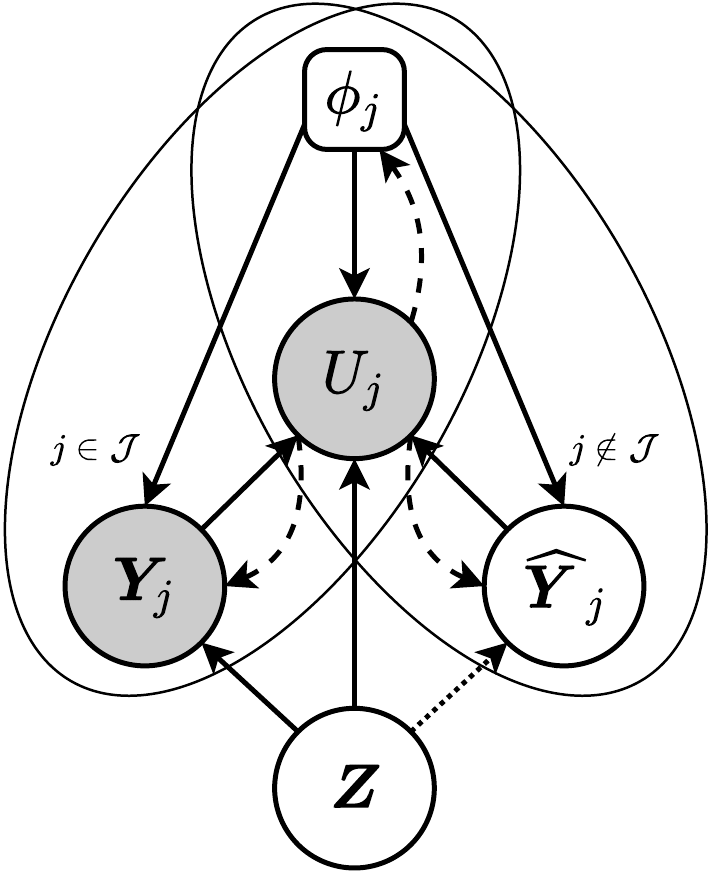}
    \caption{Extended framework.}
    \label{fig:graphical_model_b}
  \end{subfigure}
  \caption{The graphical representation of the proposed framework, where we omit replication over independent spatial locations of $\Omega$ for conciseness. Note that random variables are in circles, deterministic parameters are in rounded boxes, observed variables are shaded and ellipses indicate replication. Solid arrows indicate generation while dashed ones refer to inference procedure from a neural network. Dotted arrows indicate that the corresponding conditional probability distribution is not incorporated in posterior computation.}
\end{figure}

Given \emph{a priori} knowledge of the common anatomy through its probability distribution $P(\bm{Z})$, the Shannon's entropy $H(\bm{Z})$ measures the amount of uncertainty associated with $P(\bm{Z})$.
The reduction of this uncertainty due to the observed images is measured by the intensity-class relative entropy or mutual information
\begin{equation}\label{eq:relative_entropy}
  \begin{aligned}
    I(\bm{U},\bm{Z}) = H(\bm{U}) - H(\bm{U}\mid\bm{Z}),
  \end{aligned}
\end{equation}
where $H(\bm{U}\mid\bm{Z})$ is the conditional entropy of the observed images given the common anatomy.
Moreover, by recognizing the conditional independence assumption in \cref{eq:cond_ind} \cite{book/Cover2005}, \cref{eq:relative_entropy} can be written as
\begin{equation}\label{eq:ICMI}
  I(\bm{U},\bm{Z}) = H(\bm{U}) - \sum_{j=1}^N H(U_j\mid\bm{Z}).
\end{equation}

The intensity-class mutual information $I(\bm{U},\bm{Z})$ is maximized when $\bm{U}$ is deterministically dependent on $\bm{Z}$.
Thus, it can also be used as a similarity metric for co-registration of the observed images, provided that the distribution of the common anatomy is available.

\cref{fig:graphical_model_a} shows the graphical representation of our generic framework, where the common anatomy $\bm{Z}$ is modelled as latent variables and the spatial transformations $\bm{\phi}=\{\phi_j\}_{j=1}^N$ are incorporated as deterministic parameters.

\subsection{$\mathcal{X}$-Metric: An N-Dimensional Information\hyp{}Theoretic Framework}\label{sec:X-metric}
Unfortunately, both the total correlation and the intensity\hyp{}class mutual information requires the computation of the joint entropy $H(\bm{U})$, which is computationally prohibitive in general for $N\gg 2$.
However, the combination of these two metrics happens to cancel out this term.
Thus, we propose the following $N$-dimensional information-theoretic similarity metric by combining \cref{eq:total_correlation} and \cref{eq:ICMI}:
\begin{equation}\label{eq:X-metric}
  \mathcal{X}(\bm{U},\bm{Z}) \triangleq C(\bm{U}) + I(\bm{U},\bm{Z}) = \sum_{j=1}^N I(U_j,\bm{Z}).
\end{equation}
Apparently, the proposed metric is much more computationally favourable as it only aggregates the individual pairwise MI's between each observed image and the common anatomy.
Since this metric is a combination of total correlation and intensity-class mutual information, and is a new yet unknown metric, thus referred to as the $\mathcal{X}$-metric.

The rationale for using the proposed $\mathcal{X}$-metric in \cref{eq:X-metric} as a similarity metric to achieve co-registration is as follows:
{As {spatial correspondences among the observed images are recovered}, the reduction in uncertainty of the common anatomy due to the {observations}, measured by the mutual information $I(\bm{U},\bm{Z})$, could be substantially improved.
This could be intuitively explained as one's improved knowledge about the underlying anatomical structures with those registered images.
More formally, one can write $I(\bm{U},\bm{Z})=H(\bm{Z})-H(\bm{Z}\mid\bm{U})$.
Therefore, the uncertainty reduction may be accompanied by the sharpening of the inferred common anatomy.
That is, as $I(\bm{U},\bm{Z})$ increases, the conditional entropy $H(\bm{Z}\mid\bm{U})$ would reduce and the posterior distribution $P(\bm{Z}\mid\bm{U})$ would become more concentrated.}
Meanwhile, it is also plausible to expect that there will be a more consistent (or functional) relationship among their intensity values \cite{journal/IJCV/viola1997}, thus increasing the total correlation.
In summary, the maximization of the $\mathcal{X}$-metric implies the maximization of $I(\bm{U},\bm{Z})$ and/or $C(\bm{U})$, both of which are {nonnegative} and proper criteria for co-registration.

{Note that computation of the $\mathcal{X}$-metric requires the spatial distribution of the common anatomy, which is usually unknown \emph{a priori}.
In the following section, we develop an algorithm that applies the $\mathcal{X}$-metric to image co-registration, which estimates the spatial distribution of the common anatomy along with the {desired spatial correspondences}.
}

\subsection{$\mathcal{X}$-CoReg: Co-Registration Based on the $\mathcal{X}$-metric}\label{sec:X-CoReg}
The proposed information-theoretic co-registration algorithm based on the $\mathcal{X}$-metric, referred to as $\mathcal{X}$\hyp{}CoReg, solves the following optimization problem:
\begin{equation}\label{eq:x_opt}
  \widehat{\bm{\phi}} = \argmax_{\bm{\phi}}\max_{\bm{\alpha}} \mathcal{X}(\bm{U}[\bm{\phi}],\bm{Z}),
\end{equation}
where $\bm{U}[\bm{\phi}]$ is the warped image group by spatial transformations $\bm{\phi}$, and $\bm{\alpha}$ denotes the common-space parameters, comprising the spatial distribution $\bm{\Gamma}$ and the prior proportions $\bm{\pi}$ of the common anatomy, with $\bm{\Gamma}=(\bm{\gamma_x})_{\bm{x}\in\Omega}$ for $\bm{\gamma_x}=[\gamma_{\bm{x},1},\dots,\gamma_{\bm{x},K}]^{\intercal}\in [0,1]^K$ and $\bm{\pi}=\{\pi_k\}_{k=1}^K$ satisfying $\sum_{k=1}^K\gamma_{\bm{x},k}=\sum_{k=1}^K\pi_k=1$.
As no closed-form solution of the inner optimization can be given, we resort to coordinate ascent by alternating the following two steps: 
\begin{itemize}[leftmargin=*]
  \item Given current estimate of the spatial transformations $\bm{\phi}^{[t]}$, update the common-space parameters $\bm{\alpha}$, \emph{i.e.},
  \begin{equation}
    \bm{\alpha}^{[t+1]} = \argmax_{\bm{\alpha}}\mathcal{X}\left(\bm{U}[\bm{\phi}^{[t]}],\bm{Z}\right).
  \end{equation}
  \item Given current estimate of the common-space parameters $\bm{\alpha}^{[t+1]}$, update the spatial transformations $\bm{\phi}$ to increase the value of $\mathcal{X}$-metric, \emph{i.e.},
  \begin{equation}
    \bm{\phi}^{[t+1]} = \argmax_{\bm{\phi}}\mathcal{X}\left(\bm{U}[\bm{\phi}],\bm{Z}^{[t+1]}\right),
  \end{equation}
  where we write $\bm{Z}^{[t+1]}$ as the common anatomy with spatial distribution $\bm{\Gamma}^{[t+1]}$ for notational conciseness.
\end{itemize}

\cref{alg:X_CoReg} summarizes the above co-registration procedure.
The derivation is deferred to \cref{sec:MLE}, where its rationale is established from the perspective of maximum likelihood.
The following subsections detail its computation.

\begin{algorithm}[t]
  \KwData{The observed images $\bm{U}=\{U_j\}_{j=1}^N$\;}
  \KwIn{Number of iterations $T$, regularization coefficient $\lambda$, registration step size $\eta$\;}
  \KwOut{The estimated spatial transformations $\widehat{\bm{\phi}}=\{\widehat{\phi}_j\}_{j=1}^N$\;}
  \textbf{Initialization}: ${\phi}_j^{[0]}\triangleq\mathrm{id}$ for $j=1,\dots,N$; initialize $\bm{\pi}^{[0]}$ and $\bm{\Gamma}^{[0]}$ by \cref{eq:init}\;
  \For{$t=0,\dots,T-1$}{
    \tcc{\footnotesize Update the common-space parameters}
    $\gamma_{\bm{x},k}^{[t+1]}\triangleq\frac{\pi_k^{[t]}\prod_{j=1}^N f_{jk}^{[t]}\big(\mu_j;\phi_j^{[t]}\big)}{\sum_{k=1}^K \pi_k^{[t]}\prod_{j=1}^N f_{jk}^{[t]}\big(\mu_j;\phi_j^{[t]}\big)}$\;
    $\pi_k^{[t+1]} \triangleq \frac{\sum_{\bm{x}\in\Omega^{\bm{\phi}^{[t]}}}\gamma_{\bm{x},k}^{[t+1]}}{\sum_{k=1}^K\sum_{\bm{x}\in\Omega^{\bm{\phi}^{[t]}}}\gamma_{\bm{x},k}^{[t+1]}}$\;
    \tcc{\footnotesize Update the spatial transformations}
    $\bm{\phi}^{[t+1]} = \bm{\phi}^{[t]} - \eta\cdot \nabla \mathcal{L}\big(\bm{\phi}\mid\bm{U};\bm{\Gamma}^{[t+1]}\big)\big\rvert_{\bm{\phi}=\bm{\phi}^{[t]}}$\;
    \If{$\mathcal{L}$ converges}{
      break loop;
    }
  }
  \Return{$\widehat{\bm{\phi}}=\bm{\phi}^{[T]}$.}
  \caption{$\mathcal{X}$-CoReg}
  \label{alg:X_CoReg}
\end{algorithm}

\subsubsection{Update of the common-space parameters}\label{sec:update_template}
Given current estimate of the spatial transformations $\bm{\phi}^{[t]}$, the spatial distribution $\bm{\Gamma}=(\bm{\gamma_x})_{\bm{x}\in\Omega}$, with $\bm{\gamma_x}=[\gamma_{\bm{x},1},\dots,\gamma_{\bm{x},K}]^{\intercal}\in[0,1]^K$, are updated via computing the posterior of the common anatomy, \emph{i.e.},
\begin{equation}\label{eq:update_gamma}
  \begin{aligned}
    \gamma_{\bm{x},k}^{[t+1]}&\triangleq P\left(z_{\bm{x},k}=1\mid \bm{u_x^{\phi}}=\bm{\mu};\bm{\phi}^{[t]}\right) \\
    &= \frac{\pi_k^{[t]}\prod_{j=1}^N f_{jk}^{[t]}\big(\mu_j;\phi_j^{[t]}\big)}{\sum_{l=1}^K \pi_l^{[t]}\prod_{j=1}^N f_{jl}^{[t]}\big(\mu_j;\phi_j^{[t]}\big)},
  \end{aligned}
\end{equation}
where
\begin{equation}\label{eq:app_model}
    f_{jk}^{[t]}\big(\mu_j;\phi_j\big)\triangleq P\left(u_{\bm{x},j}^{\phi_j}=\mu_j\mid z_{\bm{x},k}=1;\phi_j,\bm{\Gamma}^{[t]}\right)
\end{equation}
describes the likelihood that the $k$-th anatomical label is observed with intensity level $\mu_j$ in the $j$-th warped image and $\bm{\mu}=(\mu_j)_{j=1}^N$.
Its form is defined in \cref{sec:density}.

Then, the prior proportions $\bm{\pi}=\{\pi_k\}_{k=1}^K$ are updated by normalizing $\bm{\Gamma}^{[t+1]}$ across the spatial domain, \emph{i.e.},
\begin{equation}\label{eq:update_pi}
  \pi_k^{[t+1]} \triangleq \frac{\sum_{\bm{x}\in\Omega^{\bm{\phi}^{[t]}}}\gamma_{\bm{x},k}^{[t+1]}}{\sum_{k=1}^K\sum_{\bm{x}\in\Omega^{\bm{\phi}^{[t]}}}\gamma_{\bm{x},k}^{[t+1]}},
\end{equation}
where the {overlap region} $\Omega^{\bm{\phi}}\triangleq\{\bm{x}\in\Omega\mid \phi_j(\bm{x})\in\Omega_j,\, \forall\,j\}$.

\subsubsection{Update of the spatial transformations}\label{sec:update_transform}
Given current estimate of the common-space parameters $\bm{\alpha}^{[t+1]}$, the spatial transformations $\bm{\phi}$ are updated by gradient descent on the loss function
\begin{equation}\label{eq:loss_function}
  \begin{aligned}
    \mathcal{L}\left(\bm{\phi}\mid\bm{U};\bm{\Gamma}^{[t+1]}\right) &\triangleq -\mathcal{X}\left(\bm{U}[\bm{\phi}],\bm{Z}^{[t+1]}\right)+\lambda\cdot R(\bm{\phi}) \\
    &= -\sum_{j=1}^N I\left(U_j\circ\phi_j,\bm{Z}^{[t+1]}\right)+\lambda\cdot R(\bm{\phi}),
  \end{aligned}
\end{equation}
where $R(\bm{\phi})$ is a regularization term with $\lambda$ its weight.
Thus, we have the following update rule for $\bm{\phi}$:
\begin{equation}\label{eq:update_phi}
  \bm{\phi}^{[t+1]} = \bm{\phi}^{[t]} - \eta\cdot \nabla \mathcal{L}\left(\bm{\phi}\mid\bm{U};\bm{\Gamma}^{[t+1]}\right)\Big\rvert_{\bm{\phi}=\bm{\phi}^{[t]}},
\end{equation}
with $\eta$ defining the step size.

The mutual information in \cref{eq:loss_function} is calculated using the formula
\begin{equation}
  \begin{aligned}
    &I\left(U_j\circ\phi_j,\bm{Z}^{[t+1]}\right) \\
    =\ & \sum_{\mu_j}\sum_{k=1}^K p_{j}^{[t+1]}(\mu_j,\!k;\phi_j) \ln\frac{p_{j}^{[t+1]}(\mu_j,\!k;\phi_j)}{p_{j}^{[t+1]}(\mu_j;\phi_j)\, p_j^{[t+1]}(k;\phi_j)},
  \end{aligned}
\end{equation}
where 
\begin{equation}\label{eq:joint_density}
  p_{j}^{[t+1]}(\mu_j,\!k;\phi_j)\triangleq P\left(u_{\bm{x},j}^{\phi_j}=\mu_j,z_{\bm{x},k}=1;\bm{\Gamma}^{[t+1]}\right),
\end{equation}
with its marginals defined by
\begin{equation}\label{eq:margin_dist}
  \begin{aligned}
    p_j^{[t+1]}(\mu_j;\phi_j) &\triangleq \sum_{k=1}^K p_{j}^{[t+1]}(\mu_j,\!k;\phi_j), \\
    p_j^{[t+1]}(k;\phi_j) &\triangleq \sum_{\mu_j} p_{j}^{[t+1]}(\mu_j,\!k;\phi_j).
  \end{aligned}
\end{equation}

\subsubsection{Density estimation}\label{sec:density}
To complete the calculation of $\mathcal{X}$-metric, it suffices to give a proper form of the conditional probability distribution $f_{jk}^{[t]}(\mu_j;\phi_j)$ in \cref{eq:app_model} and the joint probability distribution $p_j^{[t+1]}(\mu_j,k;\phi_j)$ in \cref{eq:joint_density}.
The former essentially describes the \emph{appearance model} of each anatomical structure and can be achieved by the kernel density estimator \cite{journal/media/d2006}, namely
\begin{equation}\label{eq:app_KDE}
  f_{jk}^{[t]}(\mu_j;\phi_j)\triangleq\frac{1}{\mathcal{Z}_{jk}}\sum_{\bm{x}\in\Omega_S^{\bm{\phi}}}\beta_3\left(\frac{u_{\bm{x},j}^{\phi_j}-\mu_j}{h}\right)\cdot \gamma_{\bm{x},k}^{[t]}
\end{equation}
with $\mathcal{Z}_{jk}$ the normalizing factor, $\beta_3(\cdot)$ the cubic B-spline kernel function fulfilling the partition of unity constraint \cite{journal/tip/thevenaz2000}, $h$ the bandwidth of the kernel, and $\Omega_S^{\bm{\phi}}$ an coordinate sample drawn from the overlap region.

The latter term $p_j^{[t+1]}(\mu_j,k;\phi_j)$ could be estimated via the same nonparametric approach, \emph{i.e.},
\begin{equation}
  p_j^{[t+1]}(\mu_j,k;\phi_j)\triangleq\frac{1}{\mathcal{Z}_j}\sum_{\bm{x}\in\Omega_{S'}^{\bm{\phi}}}\beta_3\left(\frac{u_{\bm{x},j}^{\phi_j}-\mu_j}{h}\right)\cdot\gamma_{\bm{x},k}^{[t+1]},
\end{equation}
where $\mathcal{Z}_j$ is the corresponding normalizing factor.

\subsubsection{Initialization}
The transformations $\bm{\phi}$ are initialized to be the identity, and the prior proportions and the spatial distribution are initialized by
\begin{equation}\label{eq:init}
  \pi_{k}^{[0]}\triangleq\frac{1}{K}, \quad \gamma_{\bm{x},k}^{[0]}\triangleq\frac{\exp(g_{\bm{x},k})}{\sum_{l=1}^K\exp(g_{\bm{x},l})},
\end{equation}
where $g_{\bm{x},k}\sim\mathcal{N}(0,1)$ for $k=1,\dots,K$ and $\forall\,\bm{x}\in\Omega$.


\subsection{Theoretical Insights from Maximum Likelihood}\label{sec:MLE}
In this section, we study the rationale of the proposed algorithm from the perspective of maximum likelihood estimation (MLE) and the expectation-maximization algorithm.

Given the generative model depicted in \cref{fig:graphical_model_a}, the log-likelihood of the observed images has the form 
\begin{equation}\label{eq:log-likelihood}
  \ell(\bm{\theta}\mid\bm{U}) = \sum_{\bm{x}\in\Omega^{\bm{\phi}}}\ln\sum_{k=1}^K\pi_k\prod_{j=1}^N {f}_{jk}\big(\mu_{\bm{x},j}^{\phi_j};\bm{\Gamma}\big),
\end{equation}
where $\mu_{\bm{x},j}^{\phi_j}$ denotes the corresponding intensity level of the resampled intensity $u_{\bm{x},j}^{\phi_j}$ and $\bm{\theta}=\bm{\alpha}\,\cup\,\bm{\phi}$ comprises all model parameters.

One standard way to find maximum likelihood solutions is by the EM algorithm \cite{journal/JRSS/dempster1977} that alternates between:

\begin{enumerate}[wide, label={(\alph*)}, labelwidth=!, labelindent=0pt]
  \item \textbf{E-step.} 
  Evaluate the posterior by
  \begin{equation*}
    q_{\bm{x},k}^{[t]}\triangleq P\left(z_{\bm{x},k}=1\mid \bm{u_x^{\phi}}=\bm{\mu};\bm{\theta}^{[t]}\right),\quad \forall\,\bm{x}\in\Omega^{\bm{\phi}^{[t]}}.
  \end{equation*}
  By definition, one can notice that for any $\bm{x}\in\Omega_{\bm{\mu}}^{\bm{\phi}}\triangleq\{\bm{x}\in\Omega^{\bm{\phi}}\mid \bm{u_x^{\phi}}=\bm{\mu}\}$, the value of $q_{\bm{x},k}$ is a constant $q_{\bm{\mu},k}$, \emph{i.e.},
  \begin{equation}\label{eq:gamma_reword}
    q_{\bm{x},k} = q_{\bm{\mu},k}, \quad\forall\,\bm{x}\in\Omega_{\bm{\mu}}^{\bm{\phi}}.
  \end{equation}
  \item \textbf{M-step.}
  Evaluate $\bm{\theta}^{[t+1]}$ given by
  \begin{equation}\label{eq:M-step}
    \bm{\theta}^{[t+1]} = \argmax_{\bm\theta}\mathcal{Q}\big(\bm{\theta}\mid\bm{\theta}^{[t]}\big),
  \end{equation}
  where ${Q}\big(\bm{\theta}\mid\bm{\theta}^{[t]}\big)$ defines the expected complete-data log-likelihood at the $t$-th step, which expands as
  \begin{equation}\label{eq:complete_log-likelihood}
    \begin{aligned}
      &\mathcal{Q}\big(\bm{\theta}\mid\bm{\theta}^{[t]}\big)\triangleq\mathbb{E}\left[\ln P(\bm{U},\bm{Z};\bm{\theta})\mid \bm{U};\bm{\theta}^{[t]}\right] \\ 
      =\ &\sum_{\bm{x}\in\Omega^{\bm{\phi}}}\sum_{k=1}^K\left[q_{\bm{x},k}^{[t]}\ln\pi_k+q_{\bm{x},k}^{[t]}\sum_{j=1}^N\ln f_{jk}\big(\mu_{\bm{x},j}^{\phi_j};\bm{\Gamma}\big)\right] \\
      =\ &\sum_{k=1}^K\sum_{\bm{\mu}}\sum_{\bm{x}\in\Omega_{\bm{\mu}}^{\bm{\phi}}}q_{\bm{x},k}^{[t]}\!\left[\ln\pi_k + \sum_{j=1}^N\ln f_{jk}(\mu_j;\phi_j,\bm{\Gamma})\right].
    \end{aligned}
  \end{equation}
  One can find that the update of $\bm{\pi}$ can be solved analytically from \cref{eq:complete_log-likelihood}, yielding
  \begin{equation}\label{eq:update_pi_Q}
    \pi_k^{[t+1]} \triangleq \frac{\sum_{\bm{x}\in\Omega^{\bm{\phi}^{[t]}}}q_{\bm{x},k}^{[t]}}{\sum_{k=1}^K\sum_{\bm{x}\in\Omega^{\bm{\phi}^{[t]}}}q_{\bm{x},k}^{[t]}}.
  \end{equation}
  Then, substituting \cref{eq:gamma_reword} into \cref{eq:complete_log-likelihood} gives
  \begin{equation}
    \!\mathcal{Q}(\bm{\theta}\mid\bm{\theta}^{[t]})\! =\! \sum_{k=1}^K\sum_{\bm{\mu}}\,\abs*{\Omega_{\bm{\mu}}^{\bm{\phi}}}\,q_{\bm{\mu},k}^{[t]}\!\left[\ln\pi_k + \sum_{j=1}^N\ln f_{jk}(\mu_j;\phi_j,\bm{\Gamma})\right],
  \end{equation}
  in which the terms involving $\bm{\phi}$ and $\bm{\Gamma}$ are rearranged into
  \begin{equation}\label{eq:S_phi}
    \mathcal{S}\big(\bm{\phi},\bm{\Gamma}\mid\bm{\theta}^{[t]}\big)\triangleq \sum_{k=1}^K\sum_{\bm{\mu}}\,\abs*{\Omega_{\bm{\mu}}^{\bm{\phi}}}\,q_{\bm{\mu},k}^{[t]}\sum_{j=1}^N\ln {f}_{jk}(\mu_j;\phi_j,\bm{\Gamma}).
  \end{equation}
  
  Note that according to the strong law of large numbers, the sample average converges almost surely to the true proportion \cite{book/Durrett2019}, namely
  \begin{equation}\label{eq:LLN}
    \frac{\abs*{\Omega_{\bm{\mu}}^{\bm{\phi}}}}{\abs*{\Omega^{\bm{\phi}}}} \stackrel{\text{a.s.}}{\longrightarrow} P^*\left(\bm{u_x^{\phi}}=\bm{\mu}\right), \quad \text{as}\ \abs*{\Omega^{\bm{\phi}}}\rightarrow\infty.
  \end{equation}
  Besides, we assume that the proposed representation can capture the true joint intensity distribution, \emph{i.e.},
  \begin{equation}\label{eq:approx}
    P^*\left(\bm{u_x^{\phi}}=\bm{\mu}\right) \approx P\left(\bm{u_x^{\phi}}=\bm{\mu}\right),\quad \forall\,\bm{x}\in\Omega^{\bm{\phi}}.
  \end{equation}
  Thus, combining \crefrange{eq:LLN}{eq:approx} yields
  \begin{equation}
    \abs*{\Omega_{\bm{\mu}}^{\bm{\phi}}}\approx \abs*{\Omega^{\bm{\phi}}}\,P\left(\bm{u_x^{\phi}}=\bm{\mu}\right).
  \end{equation}
  In addition, we allow $q_{\bm{\mu},k}$ as a function of $\bm{\Gamma}$ and $\bm{\phi}$, \emph{i.e.},
  \begin{equation}\label{eq:q_func}
    \begin{aligned}
      q^{[t]}_{\bm{\mu},k} &= q_{\bm{\mu},k}(\bm{\phi},\bm{\Gamma})\big\rvert_{\bm{\phi}=\bm{\phi}^{[t]},\bm{\Gamma}=\bm{\Gamma}^{[t]}} \\
      &= P\left(z_{\bm{x},k}=1\mid\bm{u_x^{\phi}}=\bm{\mu};{\pi}_k^{[t]},\bm{\Gamma}\right).
    \end{aligned}
  \end{equation}
  Therefore, by further denoting 
  \begin{equation}
    g_k(\bm{\mu};\bm{\phi},\bm{\Gamma})\triangleq P\big(\bm{u_x^{\phi}}=\bm{\mu},z_{\bm{x},k}=1;\pi_k^{[t]},\bm{\Gamma}\big),
  \end{equation} 
  we obtain
  \begin{equation}\label{eq:Q_JE}
    \begin{aligned}
      &\mathcal{S}(\bm{\phi},\bm{\Gamma}\mid\bm{\pi}^{[t]}) \\
      \approx\ &\abs*{\Omega^{\bm{\phi}}}\cdot\sum_{k=1}^K\sum_{\bm{\mu}}g_k(\bm{\mu};\bm{\phi},\bm{\Gamma})\sum_{j=1}^N\ln {f}_{jk}(\mu_j;\phi_j,\bm{\Gamma}) \\
      =\ & -\,\abs*{\Omega^{\bm{\phi}}}\cdot \sum_{j=1}^N H(U_j\circ\phi_j\mid\bm{Z};\bm{\Gamma}).
    \end{aligned}
  \end{equation}

  Deriving an exact update rule for $\bm{\Gamma}$ is difficult with the kernel density estimator.
  However, the following strategy was empirically found to increase the value of the right-hand side of \cref{eq:Q_JE} \emph{w.r.t.} $\bm{\Gamma}$ in general:
  \begin{equation}
    \gamma_{\bm{x},k}^{[t+1]} \triangleq q_{\bm{x},k}^{[t]},\quad \forall\,\bm{x}\in\Omega^{\bm{\phi}},
  \end{equation}
  leading to \cref{eq:update_gamma}.
  Note that this strategy represents a generalized EM scheme \cite{journal/JRSS/dempster1977}.
  Besides, \cref{eq:update_pi_Q} is thus equivalent to \cref{eq:update_pi}.
  On the other hand, the registration parameters can only be updated numerically via gradient ascent.

  Specifically, given the new spatial distribution $\bm{\Gamma}^{[t+1]}$, the terms involving $\bm{\phi}$ in the right-hand side of \cref{eq:Q_JE} are a multiple of the sum of individual conditional entropies with each observed image given the common anatomy.
  Provided that the size of the overlap region is approximately identical under spatial transformations, the objective function for updating $\bm{\phi}$ simply reduces to the sum of these conditional entropies.
  Nevertheless, to avoid a constant solution \cite{journal/IJCV/viola1997}, the mutual information $I(U_j\circ\phi_j,\bm{Z})$ is preferred so that the proposed $\mathcal{X}$-metric is optimized.

\end{enumerate}

Hence, \cref{alg:X_CoReg} is closely related to the EM algorithm for solving the MLE of the image generative model in \cref{fig:graphical_model_a}, lending strong mathematical soundness to the proposed $\mathcal{X}$-CoReg algorithm.

\subsection{Application to Groupwise Registration}\label{sec:GR}
Groupwise registration aims to spatially align multiple images simultaneously. 
The process requires a similarity metric that exploits and combines information from the entire group of images.

Delightfully, the proposed $\mathcal{X}$-metric is such a candidate.
It first summarizes images into an anatomical template (the spatial distribution of the common anatomy) representing the average shape of the group.
Then, the similarity between the average shape and each warped image is maximized to update the spatial correspondences.
The procedure is interleaved and repeated until convergence, when the anatomical correspondences among the image group are recovered.

Groupwise registration can be further categorized  according to the common space that one expects.
If the common space is implicitly assumed during co-registration, it is referred to as the \emph{unbiased} groupwise registration, as no bias is induced by assigning a certain reference image.
On the other hand, if one requires a fixed target space to which the others are simultaneously registered, then the common space is set to the target space as a reference frame and the procedure is called the \emph{group-to-reference} registration \cite{journal/media/geng2009}.

In \emph{unbiased} groupwise registration, to avoid the degeneracy that aligns images to an arbitrary coordinate space, it is suggested that one constrain the sum of all deformations to be zero \cite{conference/ISBI/bhatia2004}, \emph{i.e.},
\begin{equation}\label{eq:zero_constraint}
  \frac{1}{N}\sum_{j=1}^N \phi_j(\bm{x}) = \bm{x},\quad \forall\,\bm{x}\in\Omega,
\end{equation}
effectively registering the images to an average space.

\subsection{Application to Deep Combined Computing}\label{sec:DCC}
Beyond groupwise registration, combined computing aims to integrate registration with segmentation in the common space.
Since the proposed $\mathcal{X}$-metric measures the statistical dependency between the observed images and the common anatomy, it can naturally extend to a deep combined computing framework where registration and segmentation are performed simultaneously by neural network estimation.
Besides, the framework allows a weakly-supervised setup where the anatomical label of some images within the group is already provided.
In this section, we proceed with the description of this extended framework, the loss function to be optimized and the network architecture that integrates the system.

\subsubsection{Extended framework}
Formally, let $\{\bm{Y}_j\}_{j\in \mathcal{J}}$ be the available segmentation masks of the corresponding observed images $\{U_j\}_{j\in \mathcal{J}}$, where $\mathcal{J}$ is an index set.
Each segmentation $\bm{Y}_j$ is assumed to be a categorical random field, \emph{i.e.} $\bm{Y}_j=(\bm{y}_{j\bm{\omega}})_{\bm{\omega}\in\Omega_j}$, with $\bm{y}_{j\bm{\omega}}=[y_{j\bm{\omega},1},\dots,y_{j\bm{\omega},K}]\in\{0, 1\}^K$ its one-hot representation.
For the $j$-th image, the probability maps $\widehat{\bm{\rho}}_j$ of the segmentation $\widehat{\bm{Y}}_j$ are inferred from a neural network.

\cref{fig:graphical_model_b} presents the graphical model for the extended framework.
Specifically, given the common space anatomy $\bm{Z}$, the observed segmentation $\bm{Y}_j$ is assumed to be sampled from the conditional distribution
\begin{equation}
  P\big(y_{j\phi_j(\bm{x}),k}=1\mid z_{\bm{x},l}=1\big) = \delta_{kl}\cdot \rho_{jk}(\phi_j(\bm{x})),\quad \forall\,\bm{x}\in\Omega,
\end{equation}
where $\rho_{jk}(\phi_j(\bm{x}))\propto\exp\left[\tau\cdot D_{jk}\left(\phi_j(\bm{x})\right)\right]$, $\delta_{kl}$ is the Kronecker delta, $D_{jk}$ is the signed distance map of $\bm{Y}_j$ for label $k$, and $\tau$ controls the slope of the distance.

Then, the initial appearance model, \emph{i.e.} the conditional distribution of the observed image $U_j$ given $\bm{Y}_j$ or $\widehat{\bm{Y}}_j$ at $t=0$, is calculated by 
\begin{equation}
  \begin{aligned}
    f_{jk}^{[0]}(\mu_j) &\propto \sum_{\bm{\omega}\in\Omega_j}\beta_3\left(\frac{U_j(\bm{\omega})-\mu_j}{h}\right)\cdot \rho_{jk}(\bm{\omega}), \quad j\in\mathcal{J}, \\
    \widehat{f}_{jk}^{[0]}(\mu_j) &\propto \sum_{\bm{\omega}\in\Omega_j}\beta_3\left(\frac{U_j(\bm{\omega})-\mu_j}{h}\right)\cdot \widehat{\rho}_{jk}(\bm{\omega}), \quad j\notin\mathcal{J},
  \end{aligned}
\end{equation}
where $\widehat{{\rho}}_{jk}(\bm{\omega})\triangleq P(\widehat{y}_{j\bm{\omega},k}=1\mid U_j)$ for $k=1,\dots,K$ and $\forall\,\bm{\omega}\in\Omega_j$ are predicted by a neural network.
However, for $t\geq 1$, both $f_{jk}^{[t]}(\mu_{\bm{x},j}^{\phi_j})$ and $\widehat{f}_{jk}^{[t]}(\mu_{\bm{x},j}^{\phi_j})$ are computed in the same way as \cref{eq:app_KDE}.

Thus, the posterior distribution of the common anatomy takes the form
\begin{equation}
  q_{\bm{x},k}^{[t]}\propto{\pi_k^{[t]}\prod_{j\in\mathcal{J}}\rho_{jk}(\phi_j(\bm{x}))f_{jk}^{[t]}(\mu_{\bm{x},j}^{\phi_j})\prod_{j\notin\mathcal{J}}\widehat{f}_{jk}^{[t]}(\mu_{\bm{x},j}^{\phi_j})}
\end{equation}
for $k=1,\dots,K$ and $\forall\,\bm{x}\in\Omega$.

\subsubsection{Loss function}
To optimize the parameters of the extended framework $\bm{\Theta}=\{\bm{\phi},\widehat{\bm{\rho}}\}$, where $\widehat{\bm{\rho}}=\{\widehat{\bm{\rho}}_j\}_{j=1}^N$ with $\widehat{\bm{\rho}}_j\triangleq\bigcup_{k=1}^K\{\rho_{jk}(\bm{\omega}):\bm{\omega}\in\Omega_j\}$,
we resort to the EM algorithm and its connection to the proposed $\mathcal{X}$-metric.
Thus, the total loss function for optimizing deep combined computing has three parts:
1) a registration loss $\mathcal{L}_1(\bm{\phi})$ using the proposed $\mathcal{X}$-metric,
2) a hybrid loss $\mathcal{L}_2(\bm{\phi},\widehat{\bm{\rho}})$ that optimizes both registration and segmentation, and
3) a segmentation loss $\mathcal{L}_3(\widehat{\bm{\rho}})$ that optimizes the probability maps for both observed and unobserved segmentation masks.

Specifically, the expected complete\hyp{}data log\hyp{}likelihood of the extended framework can be arranged into two terms involving the network parameters.
One term is the cross entropy between the posterior and the appearance model.
For images with clear intensity-class correspondence, we approximate it using the proposed $\mathcal{X}$-metric, which combines with regularization to fulfil registration by the loss function
\begin{equation}
  \mathcal{L}_1(\bm{\phi})\triangleq - \mathcal{X}(\bm{U}[\bm{\phi}],\bm{Z}^{[2]}) + \lambda\cdot R(\bm{\phi}),
\end{equation}
where $\bm{Z}^{[2]}$ is the common anatomy with the spatial distribution updated by \cref{eq:update_gamma} and \cref{eq:update_pi} for $t=0,1$.
The term $R(\bm{\phi})$ is given by the aggregated bending energy with all spatial transformations \cite{journal/tmi/rueckert1999}.
Detailed explanations of the approximation can be found in Sec. 1.1 of the supplementary material.
Note that we use $\bm{Z}^{[2]}$ instead of $\bm{Z}^{[1]}$ because at $t=0$ the appearance model is calculated using the probability maps of the image anatomy rather than the spatial distribution of the common anatomy.

The other term of the complete-data log-likelihood is a hybrid loss given by the cross entropy between the posterior and the warped probability maps, \emph{i.e.},
\begin{equation}\label{eq:posterior_CE}
  \mathcal{L}_2(\bm{\phi},\widehat{\bm{\rho}}) \triangleq \sum_{j\in\mathcal{J}} H_{\bm{Z}^{[2]}}\big(\bm{Y}_j\circ\phi_j\big) + \sum_{j\notin\mathcal{J}} H_{\bm{Z}^{[2]}}\big(\widehat{\bm{Y}}_j\circ\phi_j\big),
\end{equation}
where 
\begin{equation}
  H_{\bm{Z}^{[2]}}\big({\bm{Y}}_j\circ\phi_j\big) \triangleq - \frac{1}{\abs{\Omega}}\sum_{\bm{x}\in\Omega}\sum_{k=1}^K \gamma_{\bm{x},k}^{[2]}\cdot \log {\rho}_{jk}(\phi_j(\bm{x})),
\end{equation}
and
\begin{equation}
  H_{\bm{Z}^{[2]}}\big(\widehat{\bm{Y}}_j\circ\phi_j\big) \triangleq - \frac{1}{\abs{\Omega}}\sum_{\bm{x}\in\Omega}\sum_{k=1}^K \gamma_{\bm{x},k}^{[2]}\cdot \log \widehat{\rho}_{jk}(\phi_j(\bm{x})).
\end{equation}
Note that gradient of $\mathcal{L}_2$ passes through both $\phi_j$ and $\widehat{\bm{\rho}}_j$ for $j\notin \mathcal{J}$ while only through $\phi_j$ for $j\in\mathcal{J}$.
Therefore, it can optimize both registration and segmentation.

Finally, we include an additional segmentation loss, \emph{i.e.},
\begin{equation}
  \mathcal{L}_3(\widehat{\bm{\rho}}) \triangleq -\sum_{j=1}^N I(U_j,\widehat{\bm{Y}}_j) + \sum_{j\notin\mathcal{J}}H(\widehat{\bm{Y}}_j) + \sum_{j\in\mathcal{J}}\mathcal{L}_{\text{seg}}(\bm{Y}_j,\widehat{\bm{Y}}_j),
\end{equation}
where the mutual information in the first term optimizes probability maps based on image intensities, the second term encourages the probability vector $[\widehat{\rho}_{j1}(\bm{\omega}),\dots,\widehat{\rho}_{jK}(\bm{\omega})]$ to be concentrated, and the last term measures the discrepancy between the network prediction and the ground-truth segmentation using cross entropy and Dice loss, namely
\begin{equation}
  \mathcal{L}_{\text{seg}}(\bm{Y}_j,\widehat{\bm{Y}}_j)\triangleq H_{\bm{Y}_j}\big(\widehat{\bm{Y}}_j\big) + \left[1 - \mathrm{DSC}(\bm{Y}_j,\widehat{\bm{Y}}_j)\right].
\end{equation}

Hence, the total loss function takes the form
\begin{equation}
  \mathcal{L}(\bm{\phi},\widehat{\bm{\rho}}) \triangleq \mathcal{L}_1(\bm{\phi}) + \mathcal{L}_2(\bm{\phi},\widehat{\bm{\rho}}) + \mathcal{L}_3(\widehat{\bm{\rho}}).
\end{equation}

\subsubsection{Network architecture}
\begin{figure*}[t]
  \centering
  \includegraphics[width=0.9\textwidth]{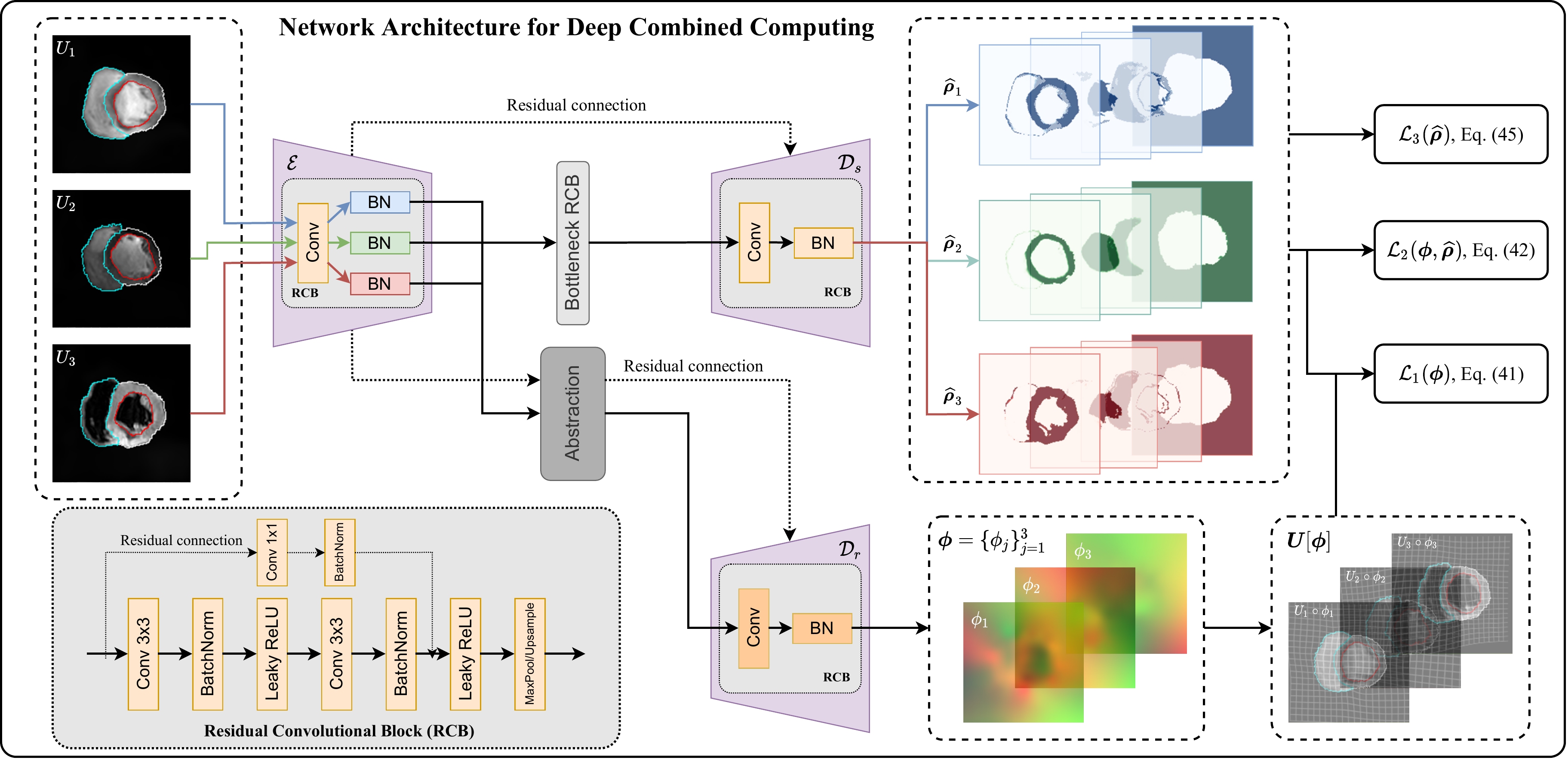}
  \caption{An example of the network architecture for deep combined computing when $N=3$. The encoder $\mathcal{E}$, the decoders $\mathcal{D}_s$ and $\mathcal{D}_r$, and the bottleneck are composed of residual convolutional blocks. Domain-specific batch normalization layers are indicated with different colours. Cardiac structures, \emph{i.e.} myocardium, left ventricle (LV) and right ventricle (RV), are rendered as contours.} 
  \label{fig:network}
\end{figure*}

\cref{fig:network} presents the network architecture for deep combined computing.
The network is composed of an encoder $\mathcal{E}$, a bottleneck, a segmentation decoder $\mathcal{D}_s$ and a registration decoder $\mathcal{D}_r$.
They {comprise} multiple levels of residual convolutional blocks (RCBs) and residual connections between the encoder and decoder \cite{journal/media/Hu2018}.
The convolutional layers of the block in the $l$-th level have $C\cdot 2^{l-1}$ feature maps, for $l=1,\dots,L_R$, with $C$ a user-defined constant.

{The task of segmentation and registration are often regarded related.
Segmentation aims to assign pixel-wise semantic labels to the input image, while registration seeks to find structural and spatial correspondences between input images.
Thus, when designing a neural network to make predictions, we believe to some extent there could be shared representations utilized by both of the tasks.
Therefore, the proposed network has a shared encoder and two separated decoders for the two tasks.}

To extract modality-invariant features, the convolutional layers of the encoder $\mathcal{E}$ are shared across modalities, while domain-specific batch normalization layers seek to disentangle structural codes from the appearances of multimodal input images \cite{conference/cvpr/chang2019}.
The segmentation decoder $\mathcal{D}_s$ is then fed with the modality-invariant features to predict the segmentation probability maps $\widehat{\bm{\rho}}$ for each image.

The extracted features of each image are also fused by an abstraction layer that computes their first and second moments \cite{conference/miccai/havaei2016}.
The fused feature maps then pass through the registration decoder $\mathcal{D}_r$ to predict the desired spatial correspondences $\bm{\phi}$ in the form of dense displacement fields.

Moreover, the network parameters for segmentation and registration are optimized alternately so that the improvement of one task can benefit the other.
{We chose to alternate training between the two branches because in \cref{eq:posterior_CE} the term $H_{\bm{Z}^{[2]}}(\widehat{\bm{Y}}_j\circ\phi_j)$ is computed using both branches.
That is, $\widehat{\bm{Y}}_j$ is predicted by the segmentation branch while $\phi_j$ is predicted by the registration branch.
Therefore, to avoid interference in two branches, like the situation where the registration branch may seek to compensate for errors in the segmentation prediction, it could be better to alternate the training for the two branches.}

\section{Experiments and Results}\label{sec:experiment}
We evaluate our method extensively on a total of five publicly available datasets, with one synthetic dataset and four clinical datasets, \emph{i.e.} BrainWeb \cite{journal/tmi/collins1998}, RIRE \cite{journal/JCAT/west1997}, MoCo \cite{journal/JBHI/pontre2016}, MS-CMR \cite{journal/tpami/Zhuang2019, journal/media/zhuang2020} and Learn2Reg-2021 \cite{journal/arxiv/hering2021}.
Our major focus is to showcase the wide applicability of our method in groupwise registration and deep combined computing.
Therefore, detailed parameter studies shall be addressed in future work.
We conducted the experiments in the following three aspects:
\begin{enumerate}[leftmargin=*]
  \item \textbf{Multimodal groupwise registration.}
  {Joint analysis of images from multiple acquisitions requires groupwise registration to a common coordinate space, where complementary information of the corresponded anatomies can be collected \cite{journal/tpami/Zhuang2019}.}
  {To perform a proof of concept for \emph{multimodal} groupwise registration, we validated the proposed method on the BrainWeb and RIRE datasets.
  The two experiments were aimed at testing the effectiveness of the $\mathcal{X}$-metric on images from various modalities and organs, as well as the situations where registration was performed with different transformation models, \emph{i.e.} nonrigid or rigid.}
  The results were compared to previously discussed groupwise similarity metrics, including the conditional template entropy (CTE) \cite{journal/media/polfliet2018}, the accumulated pairwise estimates (APE) \cite{journal/tpami/wachinger2012}, and the Gaussian mixture model (GMM) \cite{journal/tip/orchard2009}.
  \item \textbf{Spatiotemporal motion estimation.} 
  Motion estimation of dynamic medical images is essential in quantifying tissue properties from images acquired after the injection of some contrast agent. 
  {For instance, the cardiac perfusion sequences are used to calculate the myocardial perfusion reserve (MPR), \emph{i.e.} the ratio of myocardial blood flow at stress versus rest, which provides prognostic value in assessing suspected cardiovascular disease.} 
  However, motion artefacts may hamper the accuracy and robustness of image-based quantification \cite{journal/JBHI/pontre2016}. 
  We therefore investigated the performance of the proposed algorithm in correcting motion artefacts for the cardiac perfusion MR images.
  \emph{Unlike difference in imaging modalities for multimodal groupwise registration, the appearance variation of perfusion MR comes from the contrast agent, and the number of images to be registered as well as the degree of freedom with the transformation model are much larger, requiring more robust yet efficient groupwise similarity metrics.}
  The test images were from the MoCo dataset.
  Both quantitative and qualitative results were compared with alternative registration methods, including the variance of intensities (VI) \cite{journal/media/metz2011}, the congealing algorithm (CG) \cite{journal/tpami/learned2005} and the conditional template entropy (CTE) \cite{journal/media/polfliet2018}.
  \item \textbf{Deep combined computing for multimodal images.} 
  We investigate the potential of the extended framework and its integration with neural networks in realizing combined computing. 
  \emph{The task is particularly meaningful when the goal is to achieve simultaneous registration and segmentation in an end-to-end fashion.}
  The effectiveness of the proposed framework for deep combined computing was validated on the MS-CMR dataset, with various configurations of the training strategy.
  The results were compared with those from the multivariate mixture model (MvMM) \cite{journal/tpami/Zhuang2019}, a competing method for iterative combined computing.
\end{enumerate}
The algorithms were implemented in PyTorch \cite{pytorch/paszke2017} and run on an $\text{NVIDIA}^\circledR$ $\text{RTX}^\textnormal{TM}$ 3090 GPU. 
The following subsections detail the experimental design and results for each of the datasets.

\subsection{Multimodal Groupwise Registration on BrainWeb}\label{sec:BrainWeb}
\subsubsection{Experimental design}
\begin{enumerate}[wide, labelwidth=!, labelindent=0pt]
  \item \emph{\textbf{Objective.}}
  This experiment aims to demonstrate the performance of our method on multimodal \emph{nonrigid} groupwise registration for multi-sequence brain MRI from the synthetic BrainWeb dataset as a proof of concept.

  \item \emph{\textbf{Data description.}}
  The BrainWeb online database\footnote{\url{https://brainweb.bic.mni.mcgill.ca/}} provides simulated T1-, T2- and PD-weighted MRI volumes from an anatomical phantom.
  The image volumes have the physical spacing of $181\times 217\times 181\,\text{mm}^3$.
  We used the image volumes corrupted by $3\%$ noise (relative to a reference tissue) and $20\%$ intensity non-uniformity.
  The data were preprocessed by skull-stripping and the anatomical labels were remapped into foreground regions composed of cerebrospinal fluid (CSF), grey matter (GM), and white matter (WM).
  For a proof of concept, we only selected the middle slice from the axial view of each image for demonstration.
  The images were normalized by z-scoring for a fair comparison between different methods.

  \item \emph{\textbf{Recovery of initial misalignment.}}
  Since the simulated images were aligned by design, one could apply initial spatial transformations to the images and correct the misalignment through \emph{unbiased} groupwise registration.
  Readers are referred to Sec. 2.1 of the supplementary materials for details on how to generate the initial misalignments.
  
  We used multi-level isotropic FFDs as the transformation model to recover spatial correspondences from the initial misalignment. 
  The FFD spacings and the number of registration steps are described in the supplementary material.
  The deformation regularization was imposed by bending energy over the FFD meshes, with $\lambda=0.001$.
  The Adam optimizer \cite{conference/ICLR/kingma2014} was adopted to optimize the registration, with an initial step size of $\eta=0.1$.
  \item \emph{\textbf{Registration methods.}} 
  The groupwise registration methods for comparison include:
  \begin{itemize}
    \item \emph{$\mathcal{X}$-Reg-UN.} 
    This approach uses the proposed $\mathcal{X}$-metric in \cref{eq:X-metric} and the procedure in \cref{alg:X_CoReg}. 
    As no prior knowledge of the appearance model is required for the proposed algorithm, it is tagged as "\emph{UN}supervised".
    \item \emph{$\mathcal{X}$-Reg-GT.} 
    This is a variant of the proposed algorithm: The default appearance model in \cref{eq:app_KDE} is estimated in conjunction with the common space.
    Nevertheless, if the segmentation model is available for each image, one can construct the "\emph{G}round-\emph{T}ruth" appearance model directly from the intensity-class correspondences, \emph{i.e.},
    \begin{equation}\label{eq:app_model_gt}
      f_{jk}^{[t]}(\mu_j) = \frac{1}{\mathcal{Z}_{jk}}\sum_{\bm{\omega}\in\Omega_j}\beta_3\left(\frac{U_j(\bm{\omega})-\mu_j}{h}\right)\cdot y_{j\bm{\omega},k},\quad \forall\, t,
    \end{equation}
    where $\bm{Y}_j=(\bm{y}_{j\bm{\omega}})_{\bm{\omega}\in\Omega_j}$ is the categorical random field given by the segmentation of the observed image $U_j$, with $\bm{y}_{j\bm{\omega}}=[y_{j\bm{\omega},1},\dots,y_{j\bm{\omega},K}]\in\{0,1\}^K$.
    This ground-truth appearance model can be viewed as \emph{a priori} knowledge of the underlying imaging process of the common anatomy when the images were already in alignment.
    This is in the same spirit as the work of \cite{conference/miccai/leventon1998}, where prior information on the joint distribution of correctly aligned training images were used to perform pairwise registration.
    \item \emph{CTE.} 
    This method uses the conditional template entropy proposed in \cite{journal/media/polfliet2018} as the groupwise similarity metric.
    Unlike our method that computes an anatomical template as $\bm{\Gamma}$, it assumes a grey-valued template estimated by principal component analysis (PCA) from the warped images.
    \item \emph{APE.} 
    This method uses the accumulated pairwise estimates proposed in \cite{journal/tpami/wachinger2012}, computed as the sum of all pairwise mutual information from the warped images.
    Thus, it has a heavy computational burden.
    \item \emph{GMM.} 
    This method uses the MLE from a Gaussian mixture model to achieve co-registration \cite{journal/tip/orchard2009}. 
    Instead of a factorized categorical distribution assumed in our modelling, the GMM method presumes a multivariate Gaussian intensity distribution given the common anatomy.
    Note that GMM has the complexity of $\mathcal{O}(N^3)$ in general.
  \end{itemize}
  We also emphasize that the congealing (CG) algorithm is not applicable to this experiment, as a small image group ($N=3$) prevents accurate density estimation.
  Besides, for similarity metrics with kernel density estimators, we set the number of intensity levels as $L=64$ and the sample rate as 0.1.
  Moreover, four common anatomical structures were assumed for $\mathcal{X}$-CoReg and GMM, \emph{i.e.} $K=4$.
  \item \emph{\textbf{Evaluation metric.}}
  The root mean squared residual displacement error, \emph{a.k.a.} the warping index \cite{journal/tip/thevenaz2000}, was used as the evaluation metric.
  {The groupwise warping index (gWI) was calculated within the foreground regions, \emph{i.e.},
  \begin{equation}\label{eq:WI}
    \mathrm{gWI}(\bm{\phi}^{\dagger},\widehat{\bm\phi})\triangleq\frac{1}{N}\sum_{j=1}^N\sqrt{\frac{1}{\abs{\widehat{\Omega}_j^f}}\sum_{\bm{x}\in\widehat{\Omega}_j^f}\norm{\bar{r}_j(\bm{x})}_2^2},
  \end{equation}
  }
  where $\widehat{\Omega}_j^f\triangleq\{\bm{x}\in\Omega\mid \phi_j^{\dag}\circ\widehat{\phi}_j(\bm{x})\in F\}$ with $F$ the foreground region of the initial phantom and
  \begin{equation}\label{eq:res_disp}
    \bar{r}_j(\bm{x})\triangleq r_j(\bm{x})-\frac{1}{N}\sum_{j'=1}^{N}r_{j'}(\bm{x}),\quad
    r_j(\bm{x})\triangleq\phi_j^{\dag}\circ\widehat{\phi}_j(\bm{x})-\bm{x}.
  \end{equation}
  The gWI will reduce to zero when the initially misaligned images are perfectly co-registered.
\end{enumerate}

\subsubsection{Results}
\begin{table}[t]
  \small
  \centering
  \caption{{Results on the synthetic BrainWeb dataset. The table presents the mean and standard deviation of the gWIs (in millimeters) before and after groupwise registration using different methods. Statistically significant differences in gWI between $\mathcal{X}$-Reg-UN and the others, suggested by a paired $t$-test ($p<0.05$), are indicated with asterisks.}}
  {
  \begin{tabular}{L{2cm}|C{2.5cm}C{2.5cm}}
    \toprule
    Method & gWI (mm) & $p$-value \\
    \midrule
    {None} & $3.866\pm 1.882$* & $<10^{-10}$ \\
    \hdashline\noalign{\vskip 0.5ex}
    {$\mathcal{X}$-Reg-UN} & $0.476\pm 0.373$ & -- \\
    {$\mathcal{X}$-Reg-GT} & $0.485\pm 0.439$ & $0.74$ \\
    \hdashline\noalign{\vskip 0.5ex}
    {CTE \cite{journal/media/polfliet2018}} & $0.545\pm 0.509$* & $9.0\times 10^{-3}$ \\
    {APE \cite{journal/tpami/wachinger2012}} & $0.411\pm 0.298$* & $1.3\times 10^{-3}$ \\
    {GMM \cite{journal/tip/orchard2009}} & $0.500\pm 0.639$ & $0.62$ \\
    \hdashline\noalign{\vskip 0.5ex}
    {CG \cite{journal/tpami/learned2005}} & {\scriptsize N/A} & -- \\
    \bottomrule
  \end{tabular}
  }
  \label{tab:Brainweb_GWI}
\end{table}

As our algorithm falls into the category of template-based groupwise registration, we compared it with another such method called the conditional template entropy (CTE) \cite{journal/media/polfliet2018}.
We also compared our method with two more computationally demanding methods, namely the accumulated pairwise estimates (APE) \cite{journal/tpami/wachinger2012} and the Gaussian mixture model (GMM) \cite{journal/tip/orchard2009}, though their computational burden is non-negligible for a larger image group.

\cref{tab:Brainweb_GWI} presents the mean and standard deviation of the groupwise warping indices before and after co-registration using different methods.
All methods have achieved sub-millimeter accuracy on average.
In particular, the two variants of the proposed algorithm perform comparably, indicating no requirement for the ground-truth appearance model.
The two variants also perform consistently better than CTE.
This suggests that it is more reasonable to use the proposed anatomical representation of the common space than the grey-valued template computed by PCA.

The table also shows that APE works better than all template-based methods marginally, with less than 0.1 mm improvement on average.
This could be attributed to the strong modelling capacity of a pairwise MRF, if not offset by its high computational cost.
The results of GMM show noticeably larger standard deviations.
We present the box plots for post-registration warping indices in the supplementary material.
The gWIs of GMM have more outliers than the others, suggesting its inferior capture range for large deformations. 

We also demonstrate the posterior distribution of the common anatomy given by the $\mathcal{X}$-CoReg algorithm for different choices of $K$ in the supplementary material.
One can see that as $K$ increases, the proposed $\mathcal{X}$-CoReg is able to reveal the anatomical structures from the observed
images, including white matter (WM), grey matter (GM), and cerebrospinal fluid (CSF).

\subsection{Multimodal Groupwise Registration on RIRE}\label{sec:RIRE}
\subsubsection{Experimental design}
\begin{enumerate}[wide, labelwidth=!, labelindent=0pt]
  \item \emph{\textbf{Objective.}}
  This experiment aims to validate the performance of our method on multimodal \emph{rigid} groupwise registration for brain images from the RIRE dataset.

  \item \emph{\textbf{Data description.}}
  The RIRE dataset contains CT, PET and (T1w, T2w, PDw) MR scans of 18 subjects.
  Among them ground-truth correspondences of one training subject were provided by fiducial marker based rigid registration. 
  As the online evaluation system is no longer available, the evaluation was conducted by warping the registered training images with known rigid transformations, and then recovering them using different registration methods.
  \item \emph{\textbf{Generation of initial misalignment.}}
  The training subject contains 8 images of various modalities.
  These images were first registered to the CT scan using the given ground-truth rigid registration.
  Then, the registered images were resampled to physical spacing of $1\times 1\times 4 \text{mm}^3$ and cropped to size of $256\times 256\times 29$.
  A total of 50 initial random rigid transformations were generated.
  The transformations had three rotation parameters sampled from $[-15, 15]$ degrees and three translation parameters from $[-20, 20]$ mm.
  The rotation was centred at the image center and performed after translation.
  \item \emph{\textbf{Recovery of initial misalignment.}}
  To recover the initial misalignment, groupwise rigid registration was performed on the misaligned images.
  Specifically, the registration was performed in two stages.
  The three translation parameters were first optimized, followed by a full rigid registration with nine parameters: the center coordinates of rotation, the rotation angles and the translation offsets.
  To accelerate convergence, the registration was performed in four and two resolution levels for the first and second stages, respectively.
  The optimization procedure was achieved by the Adam optimizer, with the initial step sizes for translation, rotation center and rotation as 1, 1 and 0.01, respectively.
  The number of iterations for the two stages was 200 and 400, which was distributed equally among different resolution levels. 
  Moreover, the constraint in \cref{eq:zero_constraint} was imposed such that the images were registered \emph{unbiasedly}.
  \item \emph{\textbf{Registration methods.}}
  The following registration methods were compared on the RIRE dataset:
  \begin{itemize}
    \item \emph{$\mathcal{X}$-CoReg.} 
    This method uses the proposed algorithm in \cref{alg:X_CoReg}.
    \item \emph{CTE.} 
    This method uses the conditional template entropy as the groupwise similarity metric \cite{journal/media/polfliet2018}.
    \item \emph{GMM.}
    This method finds the MLE of a Gaussian mixture model to achieve groupwise registration \cite{journal/tip/orchard2009}. 
    \item \emph{APE.}
    This method uses the accumulated pairwise estimates from all pairwise mutual information as the similarity metric \cite{journal/tpami/wachinger2012}.
  \end{itemize}
  In addition, for registration methods using the kernel density estimator, 32 and 64 intensity levels were adopted for the first and second registration stage, respectively. 
  The sample rate was set as 0.1.
  Besides, the default number of common anatomical labels was assumed as 8 for $\mathcal{X}$-CoReg and GMM, \emph{i.e.} $K=8$.
  \item \emph{\textbf{Evaluation metric.}}
  We use the groupwise registration error (gRE) in eight vertices of the image volume as the evaluation metric.
  Specifically, let $V = \{\bm{v}_i\}_{i=1}^8$ be the physical coordinates of the eight vertices, with $\bm{v}_i\in\Omega\subset\mathds{R}^3$.
  The groupwise registration error is defined as the average root mean squared error of the eight vertices, \emph{i.e.},
  \begin{equation}\label{eq:gRE}
    \mathrm{gRE}(\bm{\phi}^{\dagger},\widehat{\bm{\phi}}) \triangleq \frac{1}{N}\sum_{j=1}^N\sqrt{\frac{1}{8}\sum_{i=1}^{8}\norm{\bar{e}_j(\bm{v}_i)}_2^2},
  \end{equation}
  where
  \begin{equation}
    \bar{e}_j(\bm{v}_i) \triangleq \phi_j^{\dagger}\circ\widehat{\phi}_j(\bm{v}_i)-\frac{1}{N}\sum_{j=1}^N\phi_j^{\dagger}\circ\widehat{\phi}_j(\bm{v}_i),
  \end{equation}
  with $\bm{\phi}^{\dagger}=\{\phi_j^{\dagger}\}_{j=1}^N$ the initial misalignments and $\widehat{\bm{\phi}}=\{\widehat{\phi}_j\}_{j=1}^N$ the estimated transformations.
  The gRE will reduce to zero when misaligned images are perfectly co-registered.
\end{enumerate}

\subsubsection{Results}
\begin{table}[t]
  \small
  \centering
  \caption{{Results on the RIRE dataset. The table presents the mean and standard deviation of the gREs (in millimeters) before and after groupwise registration using different methods, along with their time and GPU memory consumption ratio ($r$) compared to $\mathcal{X}$-CoReg. Statistically significant differences in gRE between $\mathcal{X}$-CoReg and the others, suggested by a paired $t$-test ($p<0.05$), are indicated with asterisks.}}
  {
  \begin{tabular}{L{1.5cm}|C{1.8cm}C{1.1cm}C{1.1cm}C{1.2cm}}
    \toprule
    Method & gRE (mm) & Time-$r$ & GPU-$r$ & $p$-value \\
    \midrule
    None & $40.65\pm 4.29$* & -- & -- & $<10^{-10}$ \\
    \hdashline\noalign{\vskip 0.5ex}
    GMM \cite{journal/tip/orchard2009} & $15.84\pm 7.78$* & 0.57 & 0.96 & $<10^{-10}$ \\
    CTE \cite{journal/media/polfliet2018} & $13.46\pm 5.04$* &0.98 & 1.01 & $<10^{-10}$ \\
    APE \cite{journal/tpami/wachinger2012} & $5.36\pm 2.60$ & 2.02 & 2.00 & $0.61$ \\
    \hdashline\noalign{\vskip 0.5ex}
    $\mathcal{X}$-CoReg & $5.52\pm 1.90$ & 1.00 & 1.00 & -- \\
    \bottomrule
  \end{tabular}
  }
  \label{tab:RIRE_gRE}
\end{table}

\cref{tab:RIRE_gRE} presents the gREs on the RIRE dataset before and after groupwise registration using different methods.
Apparently, CTE and GMM failed to achieve a good accuracy, while both APE and $\mathcal{X}$-CoReg reduced the gREs remarkably.
Note that the relatively low computational cost of GMM may be attributed to additional acceleration in PyTorch. 
However, APE required over four times the computational complexity compared to $\mathcal{X}$-CoReg as it accumulated  $\binom{8}{2}=28$ pairwise MI's, making it less efficient than our algorithm.

\begin{figure}[t]
  \centering
  \includegraphics[width=\linewidth]{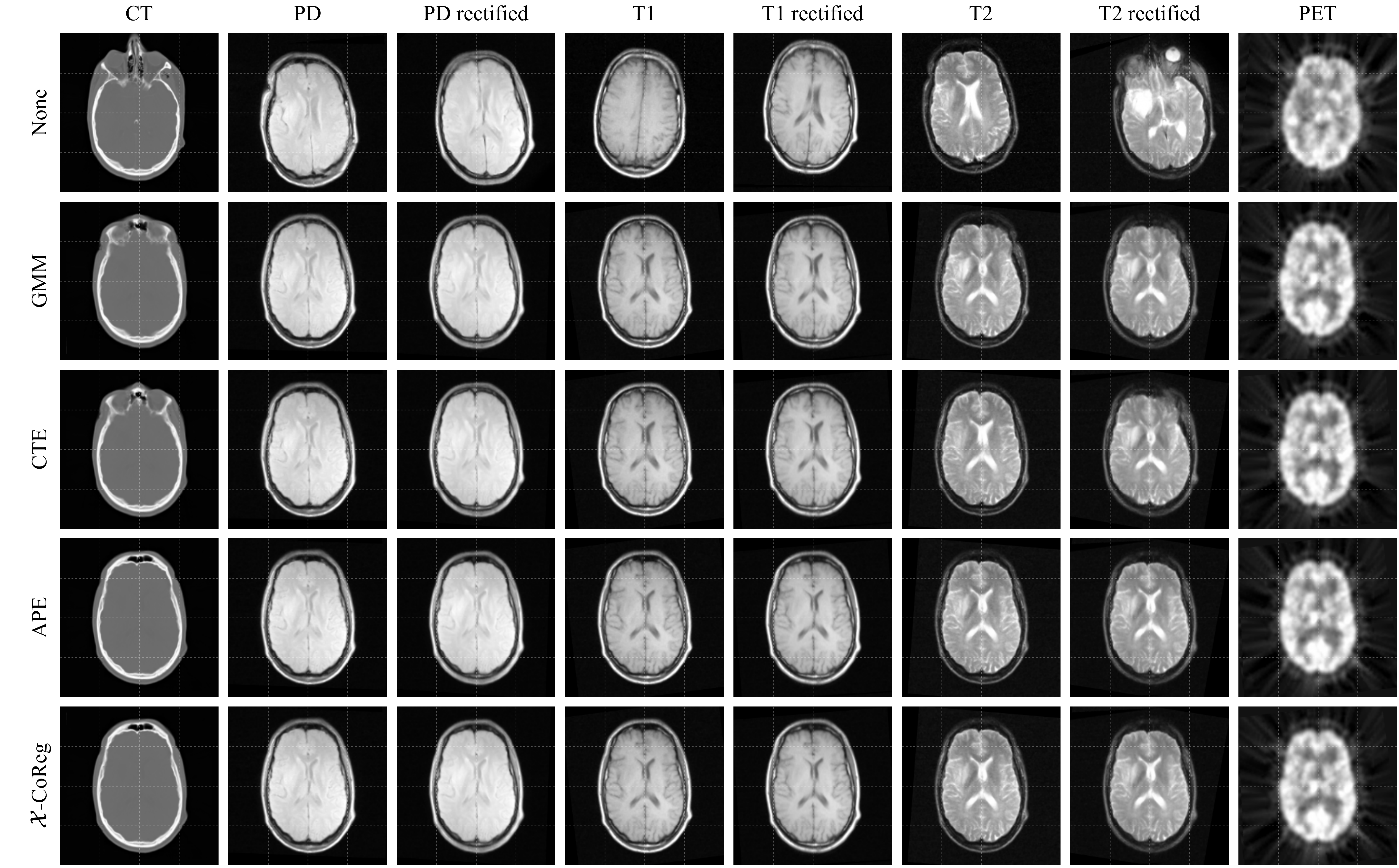}
  \caption{The registration results of an artificial misaligned RIRE training pair with median gRE before registration. 
  The middle axial slices before and after co-registration using various methods are demonstrated. 
  {Each row presents the registered images of different modalities from a certain groupwise registration method.}
  One can observe substantial initial misalignment with different images from the first row.
  Note that only APE and the proposed $\mathcal{X}$-CoReg have produced well-registered CT, T2 rectified and PET images.
  Readers are referred to the supplementary material or the online version of this paper for details.} 
  \label{fig:RIRE_rigid45}
\end{figure}

\cref{fig:RIRE_rigid45} visualizes the misaligned training pair with median gRE before registration and its registration results using different methods.
One can observe substantial initial misalignment among different images from the first row of the figure, with initial gRE of 40.179 mm.
However, neither GMM nor CTE could lead to well-registered images, especially for the CT, T2 rectified and PET scans, yielding poor gREs of 15.491 and 14.885 mm on this training pair, respectively.
On the other hand, both APE and the proposed $\mathcal{X}$-CoReg could register this training pair successfully, producing superior gREs of 6.691 and 6.102 mm, respectively.

\subsubsection{The number of common anatomical labels}
\begin{figure}
  \centering
  \includegraphics[width=0.65\linewidth]{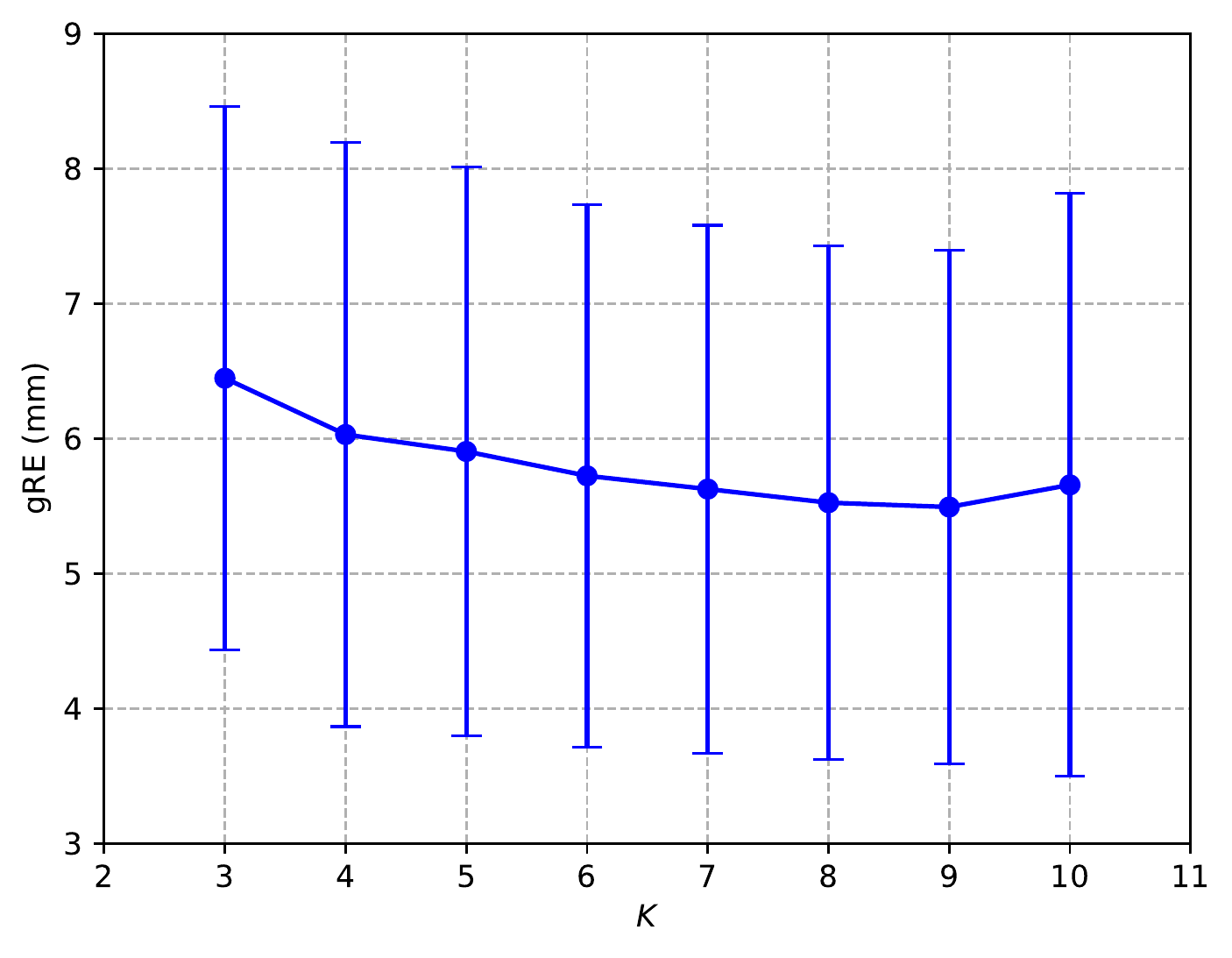}
  \caption{The influence of the number of common anatomical labels $K$ on the gRE with the RIRE dataset after registration using the proposed $\mathcal{X}$-CoReg algorithm.
  One can see improved accuracy as $K$ increases until some threshold is attained, \emph{e.g.} $K=10$.}
  \label{fig:gRE_against_K}
\end{figure}

We also studied the influence of $K$ on the registration accuracy for the $\mathcal{X}$-CoReg algorithm.
\cref{fig:gRE_against_K} plots the gRE of $\mathcal{X}$-CoReg against the number of common anatomical labels.
One can see that increasing $K$ seems to benefit the registration accuracy up to a certain level.
This could be attributed to a better clustering of the JID using $K$ that is close to the true number of common anatomical labels, leading to enhanced registration performance.
In general, the performance of our method is robust to the choice of the hyperparameter $K$.

\subsection{Spatiotemporal Motion Estimation on MoCo}\label{sec:MoCo}
\subsubsection{Experimental design}
\begin{figure*}[t]
  \centering
  \includegraphics[width=\textwidth]{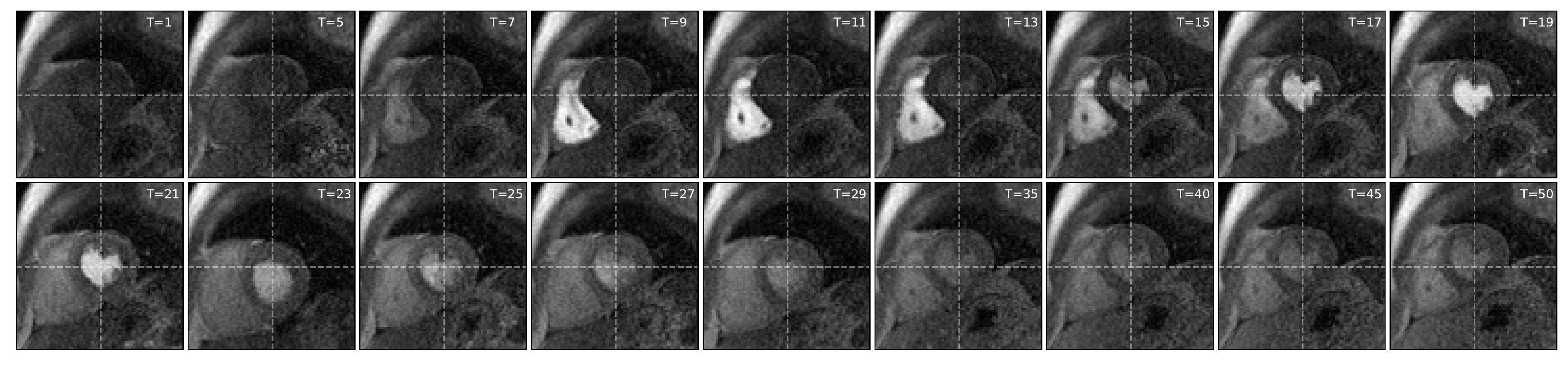}
  \caption{An example of a first-pass cardiac perfusion sequence at the adenosine induced stress condition. 
  The sequence has 50 temporal frames ($\mathrm{T}=1,\dots,50$) and only a subregion around the left ventricle myocardium is visualized. 
  At $\mathrm{T}=7$, the contrast has arrived at the right ventricle (RV) cavity. 
  At $\mathrm{T}=15$, it has reached the left ventricle (LV) cavity.
  At about $\mathrm{T}=19$, it reaches the myocardium.
  One can observe the large bulk motion at $\mathrm{T}=23,25,29$.} 
  \label{fig:MoCo_example}
\end{figure*}

\begin{enumerate}[wide, labelwidth=!, labelindent=0pt]
  \item \emph{\textbf{Objective.}}
  This experiment intends to demonstrate the performance of our method on spatiotemporal motion estimation for dynamic contrast enhanced cardiac perfusion images from the MoCo dataset.

  \item \emph{\textbf{Data description.}}
  The MoCo dataset contains mid-ventricular short-axis first-pass cardiac perfusion MR images for 10 patients at both rest and adenosine induced stress phases \cite{journal/JBHI/pontre2016}.
  Each perfusion sequence consists of 50 frames scanned over approximately 70 heartbeats, during which the T1-weighted MR imaging is used to track the uptake and washout of a contrast agent.
  The contrast agent can provide information about the myocardial blood flow, measured by the myocardial perfusion reserve (MPR).
  However, motion artefacts caused by involuntary respiration will affect the precision with image-based quantification of the MPR, as misalignment of the temporal frames can impair spatial correspondences among the myocardial regions. 
  \cref{fig:MoCo_example} shows an example of the perfusion sequence.
  \item \emph{\textbf{Motion correction procedure.}}
  To correct the motion artefacts, the groupwise registration algorithms proceeded with the following steps:
  \begin{itemize}
    \item \emph{Preprocessing.} 
    The images were preprocessed by a noise reduction filtering, z-score normalization, and the extraction of a region of interest (ROI).
    The noise reduction filtering was done by an isotropic Gaussian filter with the standard deviation as 1 pixel to avoid local optima and accelerate convergence of the registration.
    Then, to confine the information used for registration, the ROI was obtained from dilating a segmentation mask of the left ventricle (including the myocardium) on a reference frame with a circular kernel of 10-pixel radius.
    \item \emph{Rigid registration.} 
    The bulk motion constitutes a major part of the respiration-induced misalignment in the perfusion sequence \cite{journal/tpami/cordero2013}.
    To correct for this effect, a two-stage rigid registration was conducted.
    The translation was first optimized, followed by a full rigid transformation with five parameters: the center coordinates of rotation, the rotation angles and the translation offsets.
    To accelerate convergence, the registration was performed in two resolution levels for both stages.
    \item \emph{Nonrigid registration.}
    To account for the residual elastic motion artefacts caused by potentially inaccurate electrocardiogram (ECG) triggering or breath-related deformations, a nonrigid registration step was included, with FFD as the transformation model. 
    As the perfusion images were scanned at approximately the same moment in the cardiac cycle \cite{journal/JBHI/pontre2016}, a large mesh spacing of $40\times 40$ mm$^2$ were applied for the FFD control points.
    The deformation regularization was enforced by bending energy over the FFD meshes, with $\lambda=0.01$.
  \end{itemize}
  The optimization of the registration was fulfilled by the Adam optimizer, with the initial step sizes for translation, rotation center/rotation and control point displacement as 0.1, 0.1/0.001 and 0.1, respectively.
  The number of iterations for registration by translation/rigid transformation and FFD was 100/50 and 50, respectively.
  For those optimized using the multi-resolution strategy, the number of iterations were equally distributed among different resolutions.
  In addition, the zero-sum constraint on the transformations (\cref{eq:zero_constraint}) was imposed such that the groupwise registration was performed in an \emph{unbiased} fashion.
  \item \emph{\textbf{Registration methods.}}
  The following registration methods were compared on the cardiac perfusion MR images:
  \begin{itemize}
    \item \emph{$\mathcal{X}$-CoReg.} 
    This method uses the proposed algorithm with the number of intensity levels as $L=64$.
    The number of common anatomical labels was assumed as 6 to account for the varying contrast along the sequence.
    \item \emph{VI.} 
    This method uses the variance of intensities as the similarity metric \cite{journal/media/metz2011}. 
    The metric is supposed to be only applicable to \emph{monomodal} groupwise registration. 
    We include it here to verify that registering perfusion MR sequences in a monomodal manner might produce suboptimal accuracy.
    \item \emph{CG.} This method minimizes the entropy of the pixel stacks to drive the registration process, \emph{a.k.a.} the congealing algorithm \cite{journal/tpami/learned2005}.
    The implementation was based on the work of \cite{conference/miccai/balci2007}, where a Gaussian kernel $G_{\sigma}$ was used for the kernel density estimator, with the standard deviation $\sigma=0.05$ for this experiment as it produced the best accuracy.
    \item \emph{CTE.} This method uses the conditional template entropy \cite{journal/media/polfliet2018}, with the number of intensity levels as $L=64$.
  \end{itemize}
  Note that the computational cost of each method is acceptable, while a more complex algorithm is prohibitive to implement for a perfusion sequence comprising 50 frames.
  \item \emph{\textbf{Evaluation metric.}}
  To evaluate the performance of motion correction, we compute the Dice similarity coefficient (DSC) on the propagated segmentation masks for myocardium at different time points after groupwise registration.
  The segmentation masks were manually delineated at the moments where large motion was observed compared to the previous frames. 
  Roughly 18 frames were segmented for each perfusion sequence on average.
  The DSC was averaged over all pairs of the warped segmentation masks to produce the evaluation metric on one sequence.
\end{enumerate}

\subsubsection{Results}
\begin{table}[t]
  \footnotesize
  \centering
  \caption{{Results on the MoCo dataset. The table presents the mean and standard deviation of the myocardium DSC before and after motion correction with different methods and transformation models. Statistically significant differences in DSC (\%) between the proposed $\mathcal{X}$-CoReg and the others, suggested by a paired $t$-test ($p<0.05$), are indicated with asterisks. The $p$-values are reported for the results after registration with FFD.}}
  {
  \begin{tabular}{L{1.25cm}|C{1.45cm}C{1.45cm}C{1.45cm}C{1.4cm}}
    \toprule
    Method & Translation & Rigid & FFD & $p$-value \\
    \midrule
    None & $68.6\pm 11.9$* & -- & -- & $8.2\times 10^{-6}$ \\
    \hdashline\noalign{\vskip 0.5ex}
    VI \cite{journal/media/metz2011} & $75.7\pm 7.6$* & $75.1\pm 7.9$* & $70.4\pm 7.8$* & $6.6\times 10^{-5}$ \\
    CG \cite{journal/tpami/learned2005,conference/miccai/balci2007} & $76.8\pm 8.1$ & $78.8\pm 5.5$ & $79.7\pm 4.7$ & $0.60$ \\
    CTE \cite{journal/media/polfliet2018} & $75.0\pm 13.2$* & $74.9\pm 13.8$* & $74.4\pm 13.6$* & $5.5\times 10^{-3}$ \\
    \hdashline\noalign{\vskip 0.5ex}
    $\mathcal{X}$-CoReg & $78.4\pm 8.7$ & $79.2\pm 7.2$ & $80.1\pm 6.0$ & -- \\
    \bottomrule
  \end{tabular}
  }
  \label{tab:MoCo_Dice}
\end{table}


\cref{tab:MoCo_Dice} presents the DSCs on the MoCo dataset before and after motion correction using different registration methods and transformation models.
One can observe that the performance of the proposed $\mathcal{X}$-CoReg exceeds all previous methods in terms of the average DSC, with statistically significant improvement over VI and CTE for all transformation models.
Noticeably, while VI and CTE work moderately well with rigid transformation, their performance drops in nonrigid registration.
In other words, VI and CTE may only be applicable to the dynamic contrast\hyp{}enhanced perfusion sequence when the problem is to achieve global motion correction.
On the other hand, our proposed algorithm works consistently better regardless of the transformation model.

\begin{figure}[t]
  \centering
  \begin{subfigure}{\linewidth}
    \centering
    \includegraphics[width=0.6\textwidth]{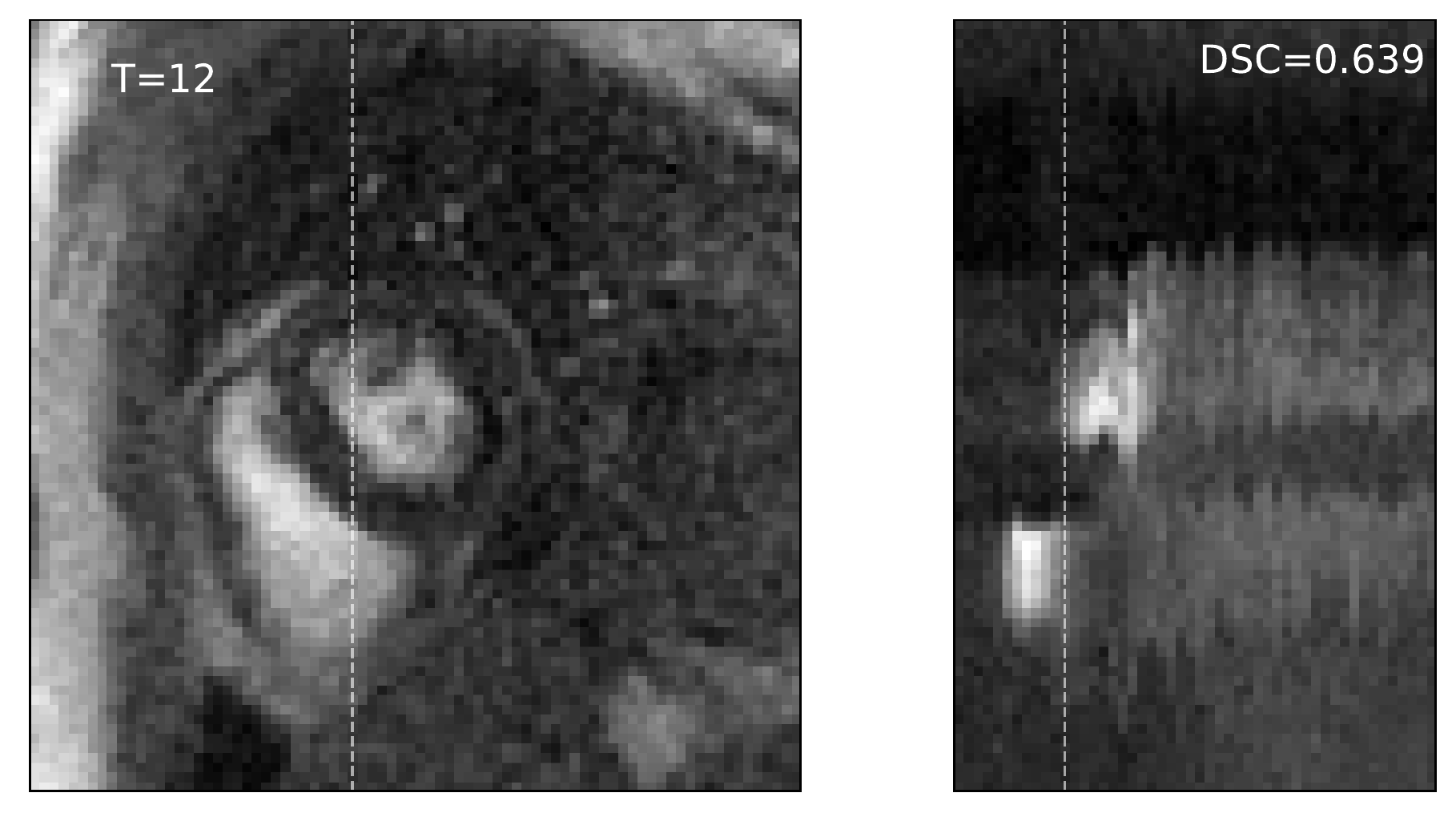}
    \caption{Before motion correction.}
  \end{subfigure}
  \hfill
  \vspace{0.2cm}
  \begin{subfigure}{\linewidth}
    \centering
    \includegraphics[width=0.9\textwidth]{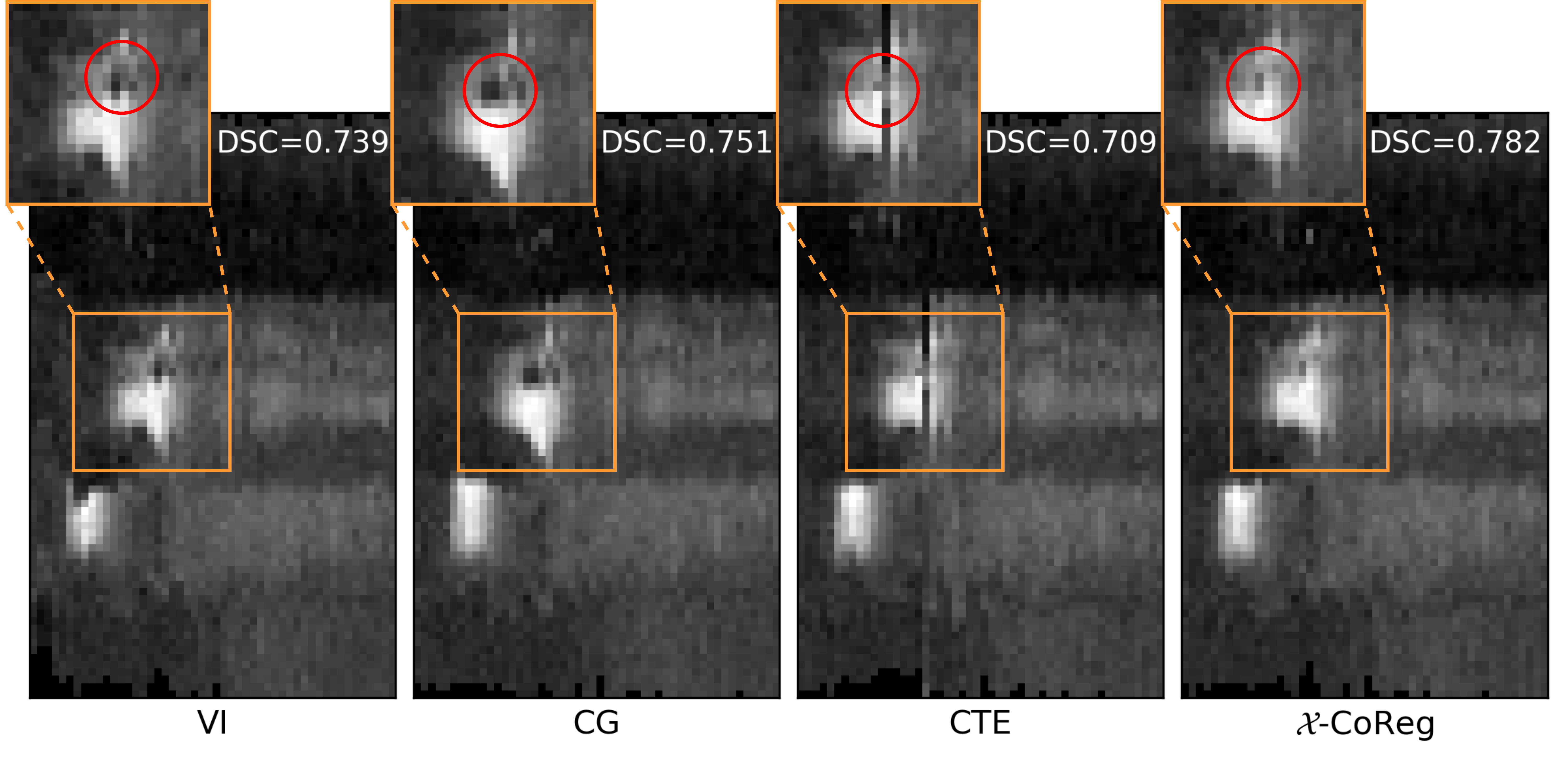}
    \caption{After motion correction.}
  \end{subfigure}
  \vspace{-0.1cm}
  \caption{The cross-sectional view along the temporal frames on the case (from the MoCo dataset) with median improvement on average using different registration methods: 
  (a) The view before motion correction.
  (b) The result after motion correction using different registration methods. 
  {One can notice that our proposed algorithm better preserves the structural correspondences,
  as indicated by the red circles.
  }
  }
  \label{fig:MoCo_04_Stress_result}
\end{figure}

\cref{fig:MoCo_04_Stress_result} visualizes the case with median improvement on average after motion correction using different registration methods.
Since the proposed algorithm is based on the assumption of the common anatomy, it better preserves the structural correspondences of the images along the temporal frames.
On the other hand, methods based only on certain assumptions with the joint intensity profile, \emph{e.g.} VI, CG and CTE, may result in disconnected anatomical structures on the registered sequence.

\subsection{Deep Combined Computing on MS-CMR}\label{sec:MSCMR}
\subsubsection{Experimental design}
\begin{enumerate}[wide, labelwidth=!, labelindent=0pt]
  \item \emph{\textbf{Objective.}}
  This experiment aims to demonstrate the performance of the extended framework on deep combined computing for multi-sequence cardiac MRI from the MS-CMR dataset.

  \item \emph{\textbf{Data description.}}
  The MS-CMR dataset provides multi-sequence cardiac MR images for 45 patients.
  Each patient was scanned by three sequences, namely the LGE (Late Gadolinium Enhanced, $j=1$), bSSFP (balanced-Steady State Free Precession, $j=2$) and T2-weighted MR ($j=3$).
  The three sequences provide complementary information of the cardiac structures.
  Namely, the LGE images can display the area of myocardium infarction, the bSSFP images present clear structural boundaries, and the T2 images can reflect the acute injury and ischemic regions \cite{journal/tpami/Zhuang2019,journal/media/zhuang2020}.
  For the purpose of demonstration, the images were preprocessed by affine co-registration, ROI cropping, background suppression and slice selection to reduce the complexity of the dataset.
  As a result, a total of 39 image slices were used for training, 15 for validation and 44 for testing.
  Besides, as the initial images are almost pre-aligned, to better demonstrate the efficacy of the proposed framework for deep combined computing, we imposed synthetic FFDs on the original images to cause additional misalignments.
  Five synthetic FFDs were generated with four different mesh spacings for each sequence.
  Moreover, the FFDs were combined among sequences during training to produce more training samples, resulting in a total of $39\times 5^3\times 4=19500$ image groups.
  On the other hand, the validation and test set comprised $15\times 5\times 4=300$ and $44\times 5\times 4=880$ image groups, respectively.

  \item \emph{\textbf{Training strategies.}}
  Four different training strategies were compared on the extended framework for deep combined computing:
  \begin{itemize}
    \item \emph{DGR.}
    This strategy uses the generic framework where no segmentation masks are observed, \emph{i.e.} $\mathcal{J}=\emptyset$.
    Thus, the problem reduces to deep groupwise registration with neural network estimation.
    Note that in this case the accuracy of the predicted segmentation results will not be evaluated, as no anatomical semantics are prescribed.
    \item \emph{DCC+AT.}
    This strategy performs deep combined computing using an atlas $\bm{Y}_4$ computed from majority voting over the training bSSFP labels.
    Therefore, it can be regarded as a special case of the extended framework for $\mathcal{J}=\{4\}$.
    \item \emph{DCC+BS.}
    This strategy used the extended framework with observed segmentation mask $\bm{Y}_2$ for bSSFP images, \emph{i.e.} $\mathcal{J}=\{2\}$.
    \item \emph{DCC+All.}
    This strategy assumes all segmentation masks of the image groups are provided, \emph{i.e.} $\mathcal{J}=\{1, 2, 3\}$.
  \end{itemize}
  Besides, for all these strategies, we used four levels of RCBs, \emph{i.e.} $L_R=4$, and the number of channels for the initial RCB was set as $C=16$.
  For computation of the $\mathcal{X}$-metric, the number of intensity levels were set as $L=32$ and the number of common anatomical labels was assumed as $K=4$ for the myocardium, left ventricle, right ventricle and the background.
  In addition, to avoid overconfidence in posterior computation caused by too concentrated probability maps, we clip the value of $\rho_{jk}(\bm{\omega})$ and $\widehat{\rho}_{jk}(\bm{\omega})$ into $[\zeta, 1-(K-1)\cdot\zeta]$, with $\zeta=0.05$ for training and $\zeta=0.01$ for testing.
  Furthermore, the constraint of \cref{eq:zero_constraint} was imposed and the bending energy with $\lambda=100$ was used for deformation regularization.

  \item \emph{\textbf{Optimization schemes.}}
  To optimize the network parameters, we used the Adam optimizer with an initial learning rate of $1 \times 10^{-3}$ for $(\mathcal{E},\mathcal{D}_s)$, and $1\times 10^{-4}$ for $\mathcal{D}_r$.
  The optimization for $(\mathcal{E},\mathcal{D}_s)$ and $\mathcal{D}_r$ were alternated every 5 training steps.
  The training process lasted for 100 epochs with a batch size of 20.
  The best model on the validation set in terms of registration accuracy was used for testing on the test data.
  \item \emph{\textbf{Competing method.}}
  We also compare our method with another iterative approach to combined computing, known as the \emph{MvMM} \cite{journal/tpami/Zhuang2019}.
  This method optimizes the likelihood function of a multivariate mixture model using the generalized EM algorithm and iterative conditional mode.
  The MvMM was initialized and regularized by the probability maps from an atlas computed in the same way as \emph{DCC+AT}.
  To re-implement the method on the MS-CMR dataset, we used FFDs with control point spacing of 40 pixels as the transformation model, with bending energy coefficient $\lambda=0.001$.
  The number of tissue subtypes were set the same as the original paper.
  The zero-sum constraint of the deformations and the value clipping of the probability maps were also imposed.
  The optimization last for 400 steps, with the initial step size as $\eta=0.1$ for the Adam optimizer.

  \item \emph{\textbf{Evaluation metrics.}}
  The performance of combined computing was evaluated in terms of both registration and segmentation.
  The registration accuracy was calculated as the DSC on the propagated segmentation masks $\{\bm{Y}_j\circ\widehat{\phi}_j\}_{j=1}^N$ with the registered images.
  The DSC was averaged over all pairs of the warped segmentation masks to produce the value.
  On the other hand, the segmentation accuracy was evaluated by the DSCs between the estimated posterior segmentation $\widehat{\bm{Z}}^{[2]}$ and each warped segmentation mask $\bm{Y}_j\circ\widehat{\phi}_j$, where  $\widehat{z}_{\bm{x},k^*}^{[2]}=1$ if and only if
  \begin{equation}
    k^*=\argmax_{k=1,\dots,K} \left\{\pi_k^{[2]}\prod_{j\in\mathcal{J}}\widehat{\rho}_{jk}(\widehat{\phi}_j(\bm{x}))\prod_{j=1}^N\widehat{f}_{jk}^{[t]}(\mu_{\bm{x},j}^{\widehat{\phi}_j})\right\}.
  \end{equation}
  Therefore, it was intended to measure errors in both registration and segmentation predictions.
\end{enumerate}

\subsubsection{Results}
\begin{table}[t]
  \scriptsize
  \centering
  \caption{Results on the MS-CMR dataset. The table presents the mean and standard deviation of the registration and segmentation DSCs (\%) for deep combined computing using different training strategies and another competing method MvMM.}
  \begin{tabular}{L{1.35cm}C{1.35cm}C{1.35cm}C{1.35cm}C{1.35cm}}
    \toprule
    \multirow{2}*{Strategy} & \multirow{2}*{Reg DSC} & \multicolumn{3}{c}{Seg DSC} \\
    \cmidrule{3-5}
    & & LGE & bSSFP & T2 \\
    \midrule
    None & $72.2\pm 10.1$ & --- & --- & --- \\
    \hdashline\noalign{\vskip 0.5ex}
    DGR & $86.2\pm 4.1$ & --- & --- & --- \\
    \hdashline\noalign{\vskip 0.5ex}
    MvMM \cite{journal/tpami/Zhuang2019} & $81.8\pm 8.7$ & $84.5\pm 9.0$ & $84.4\pm 8.5$ & $78.3\pm 14.8$ \\
    DCC+AT & $88.5\pm 3.4$ & $82.0\pm 3.8$ & $81.2\pm 4.5$ & $83.4\pm 4.1$ \\
    \hdashline\noalign{\vskip 0.5ex}
    DCC+BS & $87.6\pm 4.0$ & $85.8\pm 3.9$ & $89.9\pm 2.8$ & $86.5\pm 4.2$ \\
    DCC+All & $89.5\pm 3.5$ & $92.6\pm 2.0$ & $92.4\pm 3.1$ & $92.7\pm 3.4$ \\
    \bottomrule
  \end{tabular}
  \label{tab:MSCMR_results}
\end{table}

\begin{figure}
  \centering
  \includegraphics[width=\linewidth]{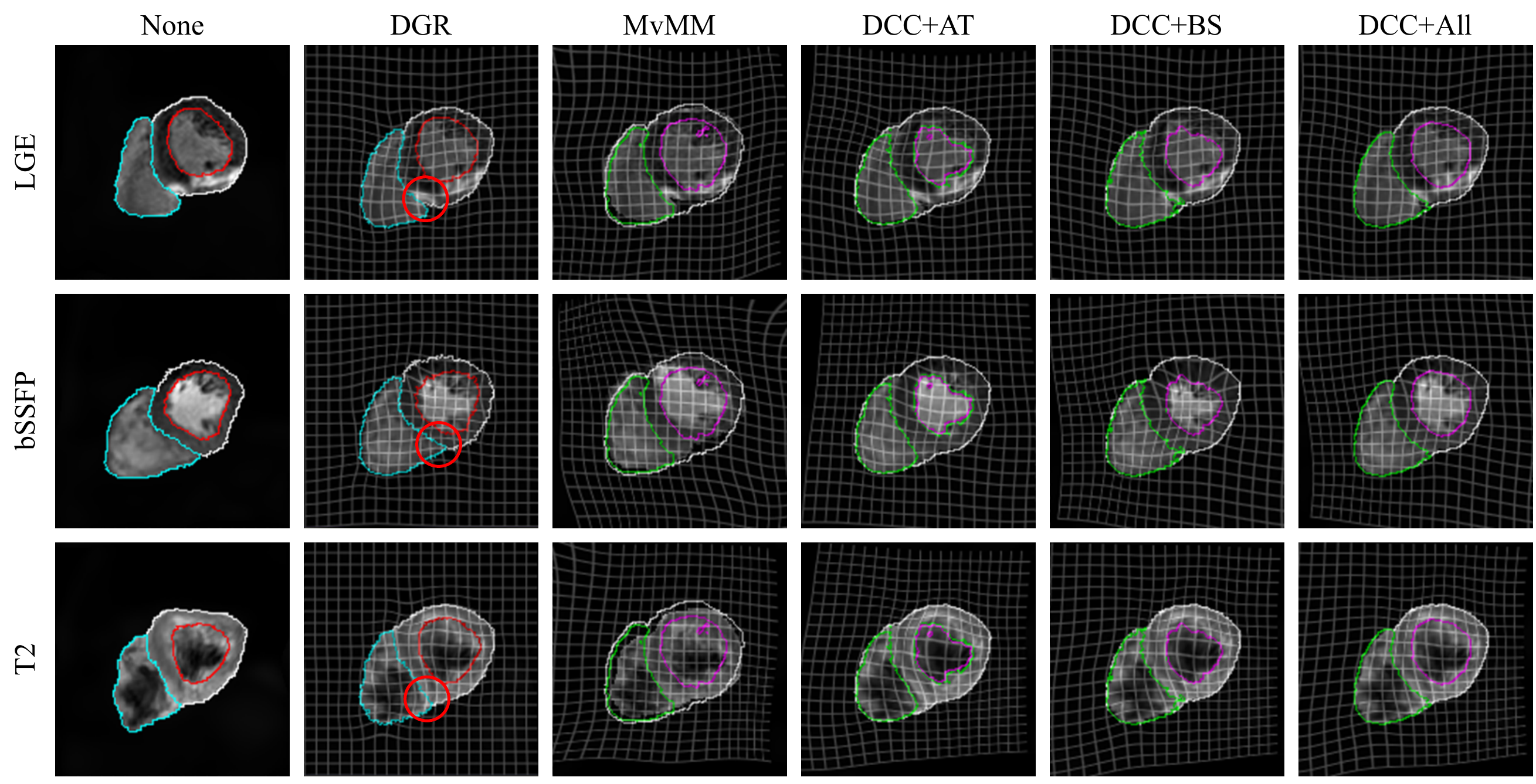}
  \caption{Results of an exemplar case from the MS-CMR dataset with median Reg DSC before co-registration.
  Ground-truth segmentation masks are rendered as contours for None and DGR, while posterior segmentation is displayed for MvMM, DCC+AT, DCC+BS and DCC+All.
  {Each column visualizes the registered images from a certain method.
  Regions with ambiguous intensity class correspondence are indicated by red circles.}
  Readers are referred to the supplementary material or the online version of this paper for details.}
  \label{fig:MSCMR_results}
\end{figure}

\cref{tab:MSCMR_results} presents the evaluation metrics on the MS-CMR dataset using MvMM and deep combined computing with different training strategies.
One can observe that both registration and segmentation accuracies of our method improve with increased supervision.
Particularly, compared to MvMM, the DCC+AT strategy performs better in registration.
\cref{fig:MSCMR_results} visualizes results from an exemplar case with median Reg DSC before co-registration.
One can see that without any supervision, the DGR strategy may produce inaccurate registration in regions with ambiguous intensity class correspondence.
On the other hand, for strategies like MvMM, DCC+AT and DCC+BS with partial or weak supervision, the posterior segmentation is satisfactory but may overplay the appearance model in posterior computation, resulting in irregular boundaries of the predicted segmentation.
Using DCC+All with full supervision can mitigate this issue and yield more accurate results.

\subsection{{Symmetric Registration on Learn2Reg-Task1}}

\begin{table}[t]
  \footnotesize
  \centering
  \caption{{Registration results on the Learn2Reg-Task1 validation set using the proposed $\mathcal{X}$-CoReg algorithm. The table presents the average DSCs (\%) before and after registration with different transformation models.}}
  {
  \begin{tabular}{L{1.3cm}|C{1.3cm}C{1.3cm}C{1.3cm}C{1.3cm}}
    \toprule
    Case ID & None & Rigid & Affine & FFD \\
    \midrule
    0012 & 24.58 & 53.71 & 68.24 & 79.99 \\
    0014 & 19.46 & 66.64 & 79.42 & 84.00 \\
    0016 & 48.49 & 76.18 & 85.32 & 87.16 \\
    \bottomrule
  \end{tabular}
  }
  \label{tab:Learn2Reg_Dice}
\end{table}

{
We also tested the proposed $\mathcal{X}$-CoReg on Task1 of the Learn2Reg-2021 challenge \cite{journal/arxiv/hering2021}.
The registration was performed symmetrically between 3 MR-CT validation pairs using rigid, affine and FFD as the transformation models.
\cref{tab:Learn2Reg_Dice} presents the registration accuracy evaluated by DSC.
One could find that our method has successfully registered these image pairs.
Please refer to Section 3.5 of the supplementary material for detailed setups and results of this experiment.
}

\section{Discussion and Conclusion}\label{sec:discussion}
In this paper, we have proposed an information\hyp{}theoretic framework that facilitates groupwise registration and deep combined computing.
The framework builds on a novel formulation of the joint intensity distribution by representing the common anatomy as latent variable and the appearance models as nonparametric estimators.
Interestingly, the proposed $\mathcal{X}$-metric can be interpreted from both the information-theoretic and the maximum-likelihood perspective:
on one hand, the $\mathcal{X}$-metric measures the shared information among the observed images through their underlying common anatomy;
on the other hand, it serves as an approximation of the objective function for finding the MLE of the image generative model with the EM algorithm.
Inspired by this connection, a co-registration algorithm named $\mathcal{X}$-CoReg is empirically found to jointly align the observed images with linear computational complexity.

\begin{figure}[t]
    \centering
    \includegraphics[width=0.9\linewidth]{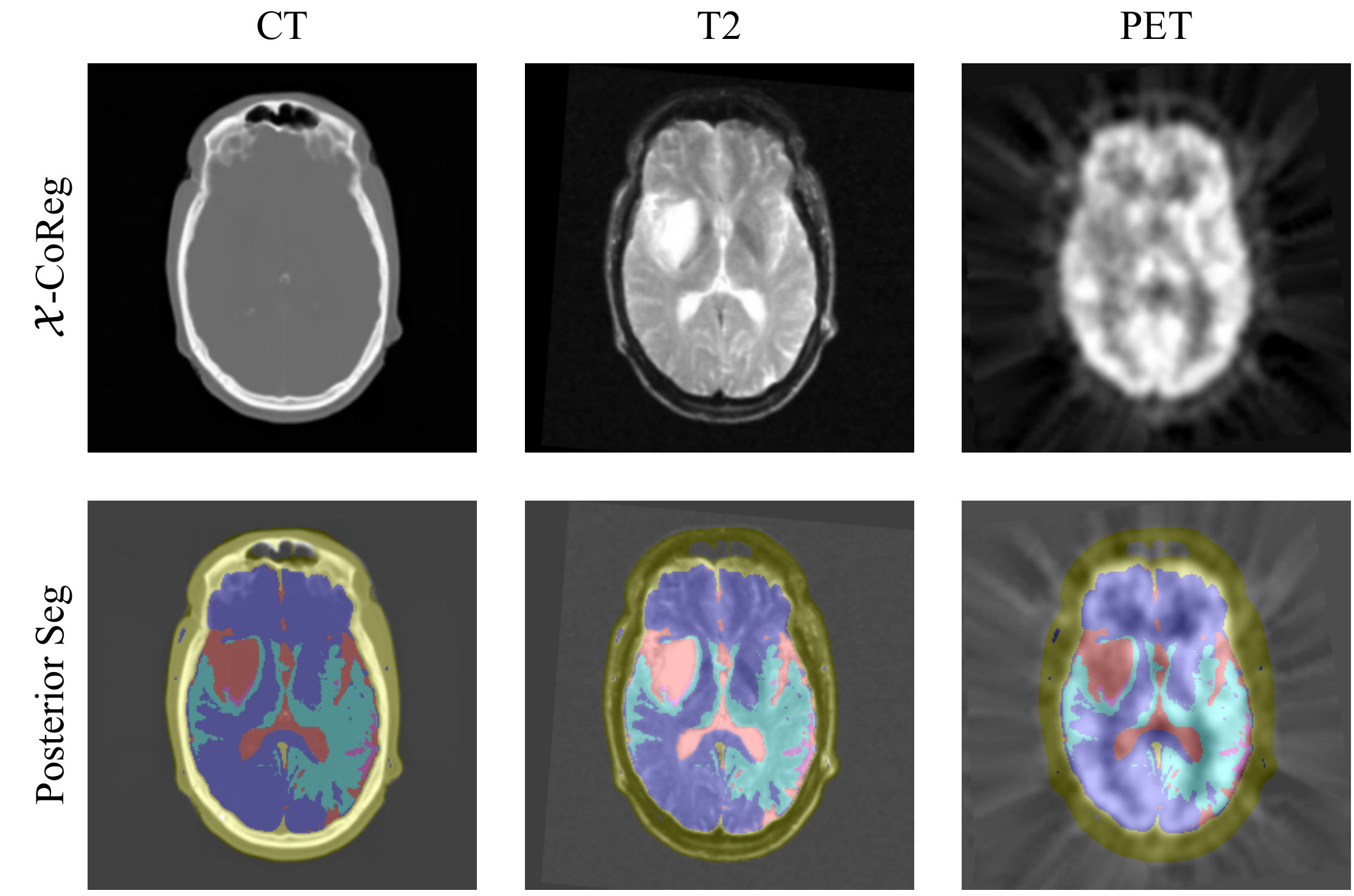}
    \caption{An example of the registered images and the posterior segmentation produced from the proposed $\mathcal{X}$-CoReg algorithm on the RIRE dataset. 
    The training pair is the same as the one used in \cref{fig:RIRE_rigid45}.
    The first row displays the registered images from the algorithm.
    One can see that different modalities exhibit complementary information of the brain anatomical structures, including the lesion visible from the T2-weighed MRI.
    The second row shows the posterior segmentation overlaid on the registered images.
    One can find that major anatomical structures are revealed from the posterior segmentation.}
    \label{fig:RIRE_posterior_seg}
  \end{figure}

To examine its applicability, we investigated a variety of tasks with the proposed framework, including multimodal groupwise registration, spatiotemporal motion estimation, and deep combined computing.
Compared to previous approaches, our method has shown great competence in its efficacy and efficiency.
Particularly, the $\mathcal{X}$-CoReg algorithm is able to register the image group even if the common anatomy is visible only from certain modalities.
\cref{fig:RIRE_posterior_seg} shows an example of the registered images produced from the algorithm on the RIRE dataset, with posterior segmentation overlaid.
One can observe that the anatomical structures manifest themselves distinctively through different imaging modalities.
Nevertheless, our method is still capable of identifying the common anatomy that interrelates the images and revealing their anatomical correspondences.
Additional details on the relationship between $K$ and the common anatomy identified from the BrainWeb images can be found in the supplementary material.

\begin{figure}[t]
  \centering
  \begin{subfigure}{\linewidth}
    \centering
    \includegraphics[width=0.6\textwidth]{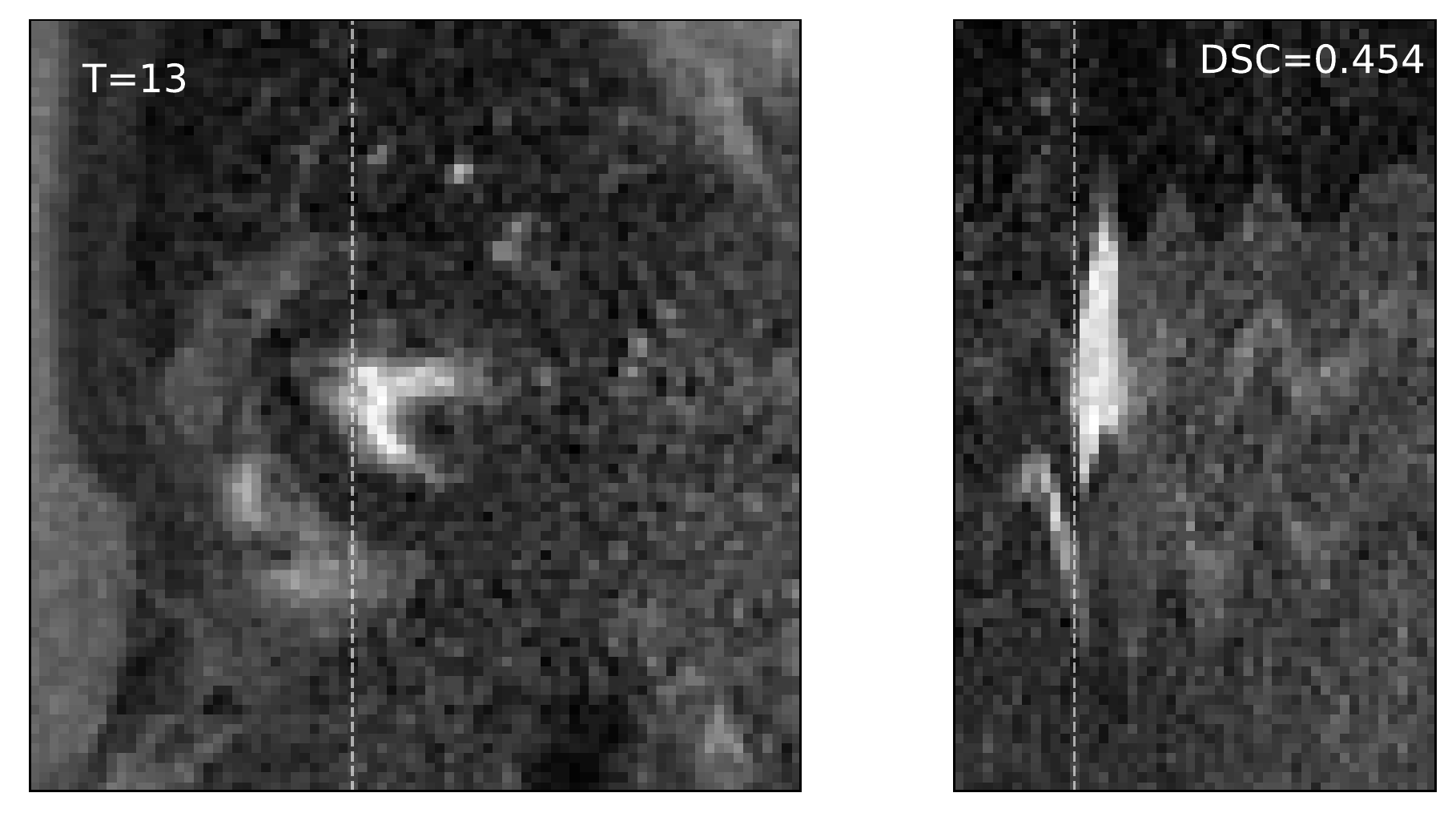}
    \caption{Before motion correction.}
  \end{subfigure}
  \hfill
  \vspace{0.1cm}
  \begin{subfigure}{\linewidth}
    \centering
    \includegraphics[width=0.9\textwidth]{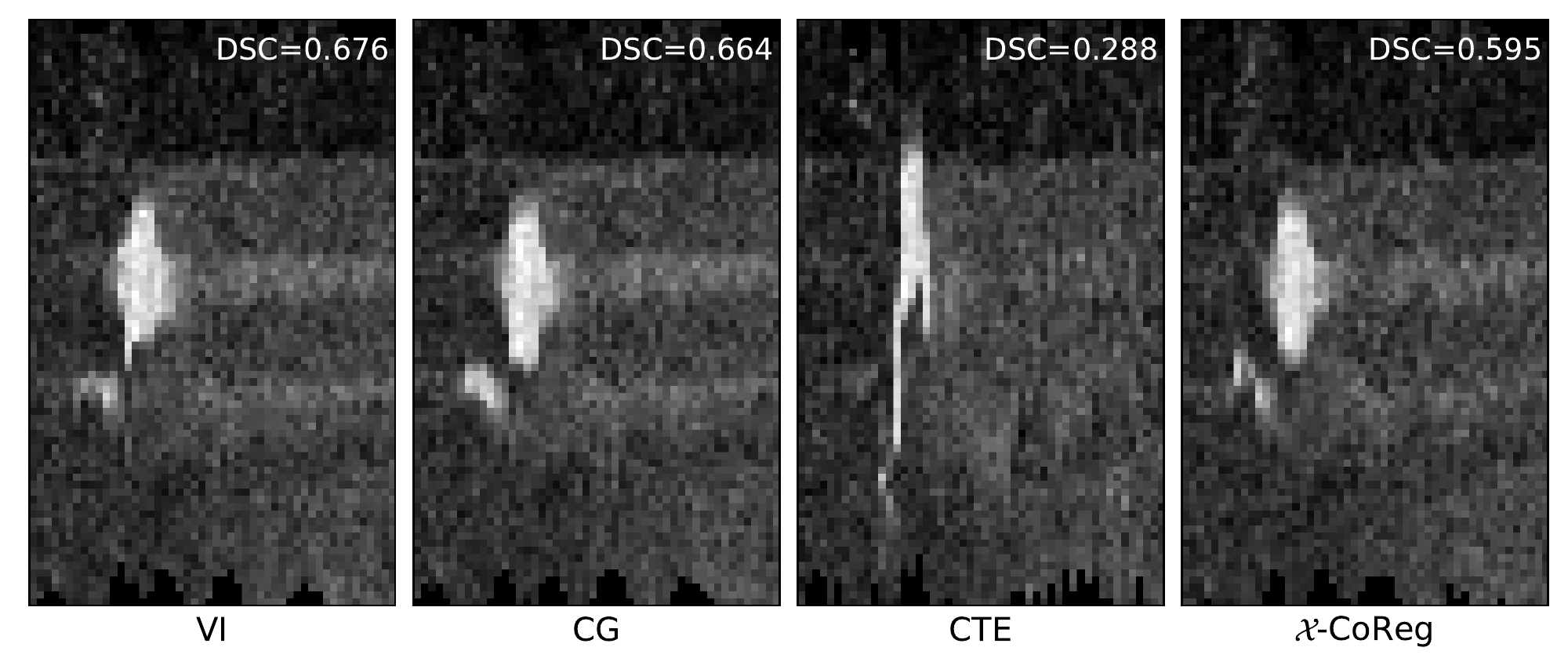}
    \caption{After motion correction.}
  \end{subfigure}
  \vspace{-0.1cm}
  \caption{The cross-sectional view along the temporal frames on the worst case (from the MoCo dataset) after alignment using the proposed $\mathcal{X}$-CoReg algorithm: 
  (a) The view before motion correction.
  One can see significant motion artefacts from the incongruent boundary of the LV myocardium.
  (b) The result after motion correction using different registration methods.}
  \label{fig:MoCo_03_Stress_result}
\end{figure}

One limitation of the proposed algorithm is the potential struggle with image artefacts, which causes the mutual information to stumble at local maxima.
An example of such effect is shown in \cref{fig:MoCo_03_Stress_result} on the MoCo dataset.
Apparently, the original sequence suffers from a low signal-to-noise ratio (SNR).
This, from the viewpoint of partition alignment, will cause the image partition resulting from histogramming to become scattered islands and not align with the anatomical boundaries, leading to suboptimal registration \cite{journal/tpami/tagare2014}.
Therefore, we will consider exploring the idea of integrating image restoration techniques into our framework to improve its robustness against these artefacts.
Analogously, several recent studies have attempted to integrate registration with image super-resolution and reconstruction to reduce error propagation and boost overall accuracy \cite{journal/tmi/lingala2014,journal/MRM/royuela2016,conference/miccai/wang2021,journal/media/corona2021}.

{Apart from the information-theoretic approach, there are also registration methods based on heuristic and hand-crafted features, as well as learning-based multimodal metrics.
For instance, \citet{journal/media/huizinga2016} designed a groupwise similarity metric particularly for quantitative MRI based on the eigenvalues of a correlation matrix.
\citet{conference/miccai/haber2006} and \citet{journal/media/heinrich2012} developed normalized gradient fields and modality-independent features for pairwise multimodal registration.
\citet{conference/miccai/simonovsky2016} proposed learning similarity metrics from registered patch pairs using convolutional neural networks, followed by \citet{journal/media/sedghi2021} who extended this notion to images only approximately registered via maximum profile likelihood. 
These formulations are nevertheless different from ours, and many of them \cite{journal/media/sedghi2021,conference/miccai/haber2006,journal/media/heinrich2012,conference/miccai/simonovsky2016} were developed for pairwise registration.
Thus, extension of those works for groupwise registration with a detailed comparison could be considered in future work.
}

In summary, the proposed framework is generic and versatile in the following three senses: 
\begin{enumerate}[leftmargin=*]
  \item It builds on the mild assumption that the observed images are interrelated through the common anatomy, given which the intensities of each image are conditionally independent. 
  The assumption is satisfied for most medical image datasets except for certain conditions where anatomical structures are only partially visible among the observed images. 
  Nevertheless, experiments on the RIRE dataset have shown that our model can still perform effectively, provided the major structures of the images can be corresponded. 
  In other words, unlike most intensity-based methods \cite{journal/tpami/learned2005,journal/tpami/wachinger2012,journal/media/polfliet2018}, the proposed framework is intended to find the intrinsic anatomical correspondences that underlie image appearances.
  \item The extended framework combines two fundamental tasks in medical image computing, registration and segmentation, the integration of which is expected to enhance their overall performance \cite{conference/miccai/xiaohua2004,journal/NI/ashburner2005,journal/NI/pohl2006}.
  Unlike previous iterative approaches, the proposed learning-based framework is able to achieve combined computing in an end-to-end fashion, exhibiting greater computational efficiency.
  Though experiments were conducted on the MS-CMR dataset for intra-subject images, the extended framework can also be applied to a more general inter-subject setup.
  In this regard, the formulation is related to multi-atlas segmentation \cite{journal/media/iglesias2015} but allows more flexibility, as expert-annotated segmentation masks are only partially needed for our method.
  \item Finally, although this paper only demonstrates the efficacy of the proposed framework in groupwise registration, motion correction and deep combined computing for complex images, we would like to emphasize that the method is not confined to these applications.
  The framework also shows potential for problems like atlas construction \cite{journal/NI/joshi2004,journal/media/lorenzen2006,journal/NI/christensen2006,journal/NI/blaiotta2018}, population analysis \cite{journal/tmi/sabuncu2009,journal/tmi/ribbens2013}, and tissue property estimation from quantitative MRI \cite{journal/media/huizinga2016}, to name a few.
\end{enumerate}
We hope to explore these implications in our future studies.







\ifCLASSOPTIONcompsoc
  \section*{Acknowledgments}
  This work was funded by the National Natural Science Foundation of China (grant no. 61971142, 62111530195 and 62011540404) and the development fund for Shanghai talents (no. 2020015).
  The authors would like to thank Fuping Wu, Hangqi Zhou, Xin Wang, Shangqi Gao and Yuncheng Zhou for their valuable comments and proofreading of the manuscript.
\else
  \section*{Acknowledgment}
\fi

\begin{scriptsize}
  \bibliographystyle{IEEEtranN}
  \bibliography{main}
\end{scriptsize}

%

\begin{IEEEbiography}[{\includegraphics[width=1in,height=1.25in,clip,keepaspectratio]{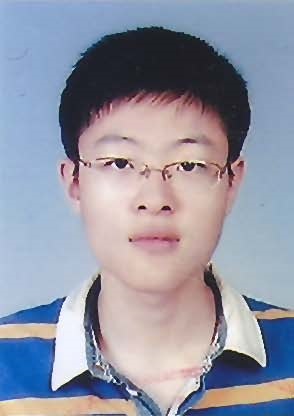}}]{Xinzhe Luo}
  is a Ph.D. student in statistics at the School of Data Science, Fudan University, advised by Prof. Xiahai Zhuang.
  He obtained his B.S. degree in information and computational science from the School of Data Science, Fudan University in 2019. 
  His research interests include medical image computing, image registration, and multivariate image analysis.
\end{IEEEbiography}

\begin{IEEEbiography}[{\includegraphics[width=1in,height=1.25in,clip,keepaspectratio]{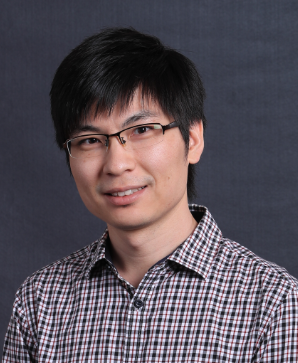}}]{Xiahai Zhuang}
  is a professor with the School of Data Science, Fudan University. 
  He graduated from the department of computer science, Tianjin University, received the MS degree in computer science from Shanghai Jiao Tong University, and obtained the doctorate degree from University College London. 
  His research interests include medical image analysis, image processing, and computer vision. 
  His works have been nominated twice for the MICCAl Young Scientist Awards (2008,2012).
\end{IEEEbiography}








\end{document}